\pdfoutput=1

\documentclass[11pt]{article}

\usepackage{ACL2023}

\usepackage{times}
\usepackage{latexsym}
\usepackage[T1]{fontenc}
\usepackage[utf8]{inputenc}
\usepackage{microtype}
\usepackage{inconsolata}

\usepackage{nicematrix}
\usepackage{url}            
\usepackage{amsfonts}       
\usepackage{xcolor,colortbl} 

\usepackage{graphicx}
\usepackage{multirow}
\usepackage{pifont}
\usepackage{amsmath}
\usepackage{subcaption} 
\usepackage{booktabs}
\usepackage{calc}
\usepackage{enumitem}

\usepackage{longtable}
\usepackage{siunitx}
\usepackage{tcolorbox}

\definecolor{headergray}{rgb}{0.86, 0.85, 0.89}
\definecolor{rowgray}{rgb}{0.95, 0.95, 0.95}
\definecolor{eqagray}{rgb}{0.90, 0.9, 0.9}

\newcommand{\accBetter}[1]{$_{\color{red}\uparrow #1}$}
\newcommand{\accWorse}[1]{$_{\color{blue}\downarrow #1}$}
\newcommand{\rmseBetter}[1]{$_{\color{red}\downarrow #1}$}
\newcommand{\rmseWorse}[1]{$_{\color{blue}\uparrow #1}$}

\definecolor{myTemporalColor}{HTML}{264653}
\definecolor{mySpatialColor}{HTML}{2A9D8F}
\definecolor{myReasoningColor}{HTML}{E9C46A}

\title{\texttt{EQA-RM}: A Generative Embodied Reward Model with Test-time Scaling}

\author{
  \textbf{Yuhang Chen}\textsuperscript{1} \quad 
  Zhen Tan\textsuperscript{2} \quad          
  \textbf{Tianlong Chen}\textsuperscript{1} \\  
  \textsuperscript{1}The University of North Carolina at Chapel Hill \quad
  \textsuperscript{2}Arizona State University \\
  {\tt \{yuhang, tianlong\}@cs.unc.edu} \quad
  {\tt ztan36@asu.edu}
}

\begin{document}
\maketitle
\begin{abstract}
 Reward Models (RMs), vital for large model alignment, are underexplored for complex embodied tasks like Embodied Question Answering (EQA) where nuanced evaluation of agents' spatial, temporal, and logical understanding is critical yet not considerred by generic approaches. We introduce \texttt{EQA-RM}, a novel generative multimodal reward model specifically architected for EQA, trained via our innovative \textbf{Contrastive Group Relative Policy Optimization} (\texttt{C-GRPO}) strategy to learn fine-grained behavioral distinctions. The generative nature of \texttt{EQA-RM} provides interpretable, structured reward feedback (beyond simple scalars), uniquely enabling test-time scaling to dynamically adjust evaluation granularity, from concise scores to detailed critiques of reasoning and grounding, at inference without retraining. Concurrently, we introduce \textsc{EQARewardBench}, a new benchmark built on OpenEQA for standardized EQA reward model assessment. Demonstrating high sample efficiency, EQA-RM (fine-tuning Qwen2-VL-2B-Instruct) achieves \textbf{61.9\%} accuracy on EQA-RM-Bench with only \textbf{700} samples, outperforming strong proprietary baselines, including Gemini-2.5-Flash, GPT-4o, Claude-3.5-Haiku, and open-sourced state-of-the-art models such as RoVRM and VisualPRM. The code and dataset can be found here: \url{https://github.com/UNITES-Lab/EQA-RM}.
\end{abstract}

\section{Introduction}

Reward Models (RMs) have emerged as a cornerstone technique in qualifying the quality of output or actions of Large Language Models. Desired RMs provide critical signals for refining model behaviors and enhancing performance, often through reinforcement learning or selection strategies~\cite{InstructGPT, snell2024scaling}.
Existing generic Reward Models~\cite{zhao2025genprm,SPCT_2025} often designed for static inputs or simple outcomes, are proved ill-equipped for dynamic and interactive domains. Their limitations is amplified in complex tasks such as Embodied Question Answering (EQA). EQA requires agents to perceive, interact, and reason through sequences of multimodal observations and actions to answer questions within 3D environments~\cite{majumdar2024openeqa,yu2019multi}. Evaluating the resulting EQA trajectories necessitates a nuanced assessment of the agent's reasoning coherence, action appropriateness, and how well language is grounded within the environment. Yet, existing RMs are usually trained via generic next-token prediction~\cite{zhao2025genprm} or classification~\cite{faal2023reward}, thus incapable to capture the spatio-temporal and logical dependencies inherent in embodied tasks. This emphasizes the urgent need for specialized mechanisms to accurately assess EQA's multifaceted success indicators.

\begin{figure*}[t]
    \centering

    \includegraphics[width=\linewidth]{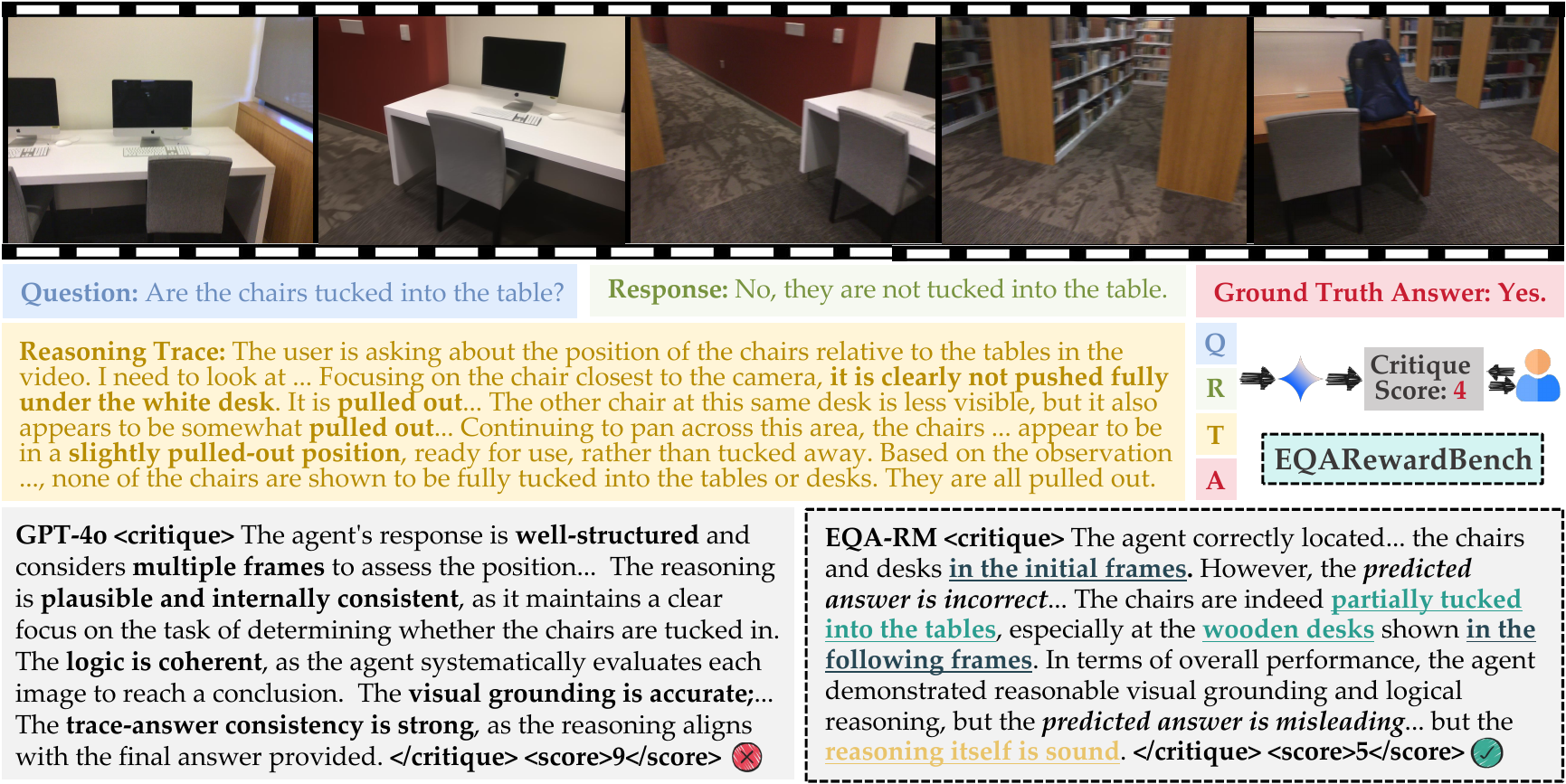}
    \vspace{-8mm}
    \caption{\small{\textsc{EQARewardBench} enables nuanced evaluation of complex EQA trajectories involving spatio-temporal reasoning. An agent's reasoning process and response to a question are assessed through detailed critiques and scores provided by reward models like our \texttt{EQA-RM} and GPT-4o. Within its critiques, \texttt{EQA-RM} explicitly assesses \textcolor{myTemporalColor}{\underline{temporal dependencies}} , \textcolor{mySpatialColor}{\underline{spatial grounding}} , and \textcolor{myReasoningColor}{\underline{reasoning coherence}} , showcasing better results than GPT-4o.}}
    \label{fig:teaser}
    \vspace{-4mm}
\end{figure*}

To address this, we propose \texttt{EQA-RM}, a novel multimodal reward model tailored for evaluating EQA trajectories. Inspired by~\citet{zhao2025genprm}, \texttt{EQA-RM} is designed as a generative reward model (GenRM). \texttt{EQA-RM} thus is able to produce not only scalar rewards but also explicit reasoning for its assessments. Such generative capability enhances interpretability and facilitate test-time scaling~\cite{snell2024scaling} for performance improvements. To effectively generate these nuanced evaluations for EQA, the \texttt{EQA-RM} is trained for enhanced spatial, temporal, and reasoning processing capabilities, enabling it to handle the unique multimodal data streams inherent in EQA tasks.

We train \texttt{EQA-RM} using an efficient two-stage process. The \underline{first stage} involves standard Rejective Finetuning (RFT)~\cite{yuan2023scaling} to teach the model the desired output format, which includes a textual critique and a scalar score. However, we observed that relying solely on RFT tends to make the model proficient at generating outputs in the learned format without necessarily achieving a deep understanding of crucial trajectory elements. This potential for supervised finetuning to primarily capture style rather than substance in preference learning has been discussed in prior work~\cite{chen2025sft, SPCT_2025}. To address this intrinsic limitation, we employ a reinforcement learning approach for the \underline{second stage}, which helps the model internalize output quality by optimizing for rewards, overcoming SFT's stylistic memorization. Specifically, we design a novel Contrastive Group Relative Policy Optimization (\texttt{C-GRPO}) strategy. \texttt{C-GRPO} utilizes rule-based contrastive rewards derived from targeted data augmentations. It optimizes \texttt{EQA-RM}'s evaluative acuity by training it to distinguish policy outputs based on their evaluation under original, coherent contexts versus synthetically perturbed ones. We design the following augmentations for perturbations:
\ding{182} trajectories with shuffled video frames, \ding{183} frames with randomly masked spatial regions, and \ding{184} sequences with jumbled reasoning steps. Essentially, \texttt{EQA-RM} earns a positive reward only when its score assigned under original, unperturbed conditions is more accurate (i.e., closer to a preference score) than its score assigned under the corresponding perturbed conditions. By learning this differentiated accuracy in its scoring relative to preference score, \texttt{C-GRPO} compels \texttt{EQA-RM} to internalize the importance of temporal order, fine-grained spatial details, and coherent logical flow. This cultivates a robust and discerning evaluative capability for embodied tasks.

On the other hand, the EQA domain lacks standardized benchmarks for rigorously evaluating and comparing reward models. Current EQA task benchmarks focus on coarse success metrics rather than the fine-grained trajectory quality crucial for RM development. To bridge this gap, we introduce \textsc{EQARewardBench}. Built upon the OpenEQA dataset, \textsc{EQARewardBench} features embodied episode memory videos from two types of household environments: HM3D~\cite{} and ScanNet~\cite{}. From the original question-answer pairs, we construct more comprehensive question-response-reasoning trace triplets. The benchmark includes 1,546 test instances designed to evaluate reward models across eight distinct aspects of trajectory quality (e.g., correctness, grounding, efficiency). \textsc{EQARewardBench} thus provides a standardized platform for the rigorous and comparable assessment of reward models on EQA tasks.

Empirical evaluations demonstrate the capabilities of \texttt{EQA-RM}. Based on Qwen2-VL-2B-Instruct, \texttt{EQA-RM} substantially improves upon its base model and outperforms existing open-source visual reward models on \textsc{EQARewardBench}. \texttt{EQA-RM} also shows test-time scalability. By increasing evaluative computations at inference, its accuracy on \textsc{EQARewardBench} is boosted from 42.47\% to 61.86\%. This enhanced performance allows \texttt{EQA-RM} to surpass leading large commercial models in accuracy on our benchmark.

In conclusion, our core contributions are: (\textbf{\textit{i}}) We propose \texttt{EQA-RM}, a generative multimodal reward model with enhanced spatial, temporal, and reasoning capabilities tailored for EQA;
(\textbf{\textit{ii}}) We introduce a EQA Reward Model Benchmark, \textsc{EQARewardBench}, the first dedicated benchmark for standardized evaluation of reward models for EQA.
(\textbf{\textit{iii}}) Extensive experiments have proved the improvement of our method for test time scaling.
    
\section{Related Work}

\noindent\textbf{Generative Reward Models.}
Reward Models are crucial for guiding LLMs outputs~\cite{ouyang2022training, christiano2017deep, ziegler2019fine}. To overcome the limitations of scalar rewards and provide deeper evaluative insights, Generative Reward Models emerged and produced rich, interpretable textual feedback like critiques or explanations~\cite{zhang2024generativeverifiers, mahan2024generative, zheng2023judging}. LLM-as-a-judge method accommodates pairwise critiques to evaluate LLMs outputs and enhances interpretability~\cite{zheng2023judging}. In multimodal domains, GenRMs enable finer-grained supervision through techniques such as step-wise reasoning assessment~\cite{wang2025visualprm}, self-critique~\cite{yu2025criticrm}, or Chain-of-Thought style evaluations\cite{zhao2025genprm}, enhancing the performance of Multimodal LLMs.

\noindent\textbf{Embodied Question Answering.}
EQA challenges agents to perceive, navigate, and interact within 3D environments to answer questions, integrating vision, language, reasoning, and planning~\cite{das2018embodied, gordon2018iqa}. While datasets like OpenEQA offer rich scenarios for this task~\cite{wijmans2019embodied, kolve2017ai2thor, majumdar2024openeqa} and EQA methodologies have advanced towards end-to-end foundation models~\cite{ahn2022saycan, driess2023palm}, current EQA evaluation predominantly assesses only final answer correctness~\cite{majumdar2024openeqa, das2018embodied}. This common practice overlooks crucial trajectory qualities such as reasoning coherence and spatio-temporal understanding~\cite{chen2025reward}, creating a significant gap in reward modeling for comprehensive EQA assessment, motivating our development of the specialized \texttt{EQA-RM}.

\noindent\textbf{Rule-based RL for LLMs.}
Reinforcement learning is increasingly used to refine LLMs for enhanced alignment and reasoning capabilities, moving beyond standard SFT~\cite{schulman2017proximal, ouyang2022training, zhai2024fine}. A key direction involves rule-based RL, which employs systematic or synthetic feedback for improved efficiency and targeted behavior control, with algorithms like Group Relative Policy Optimization (GRPO)~\cite{shao2402deepseekmath} notably using relative comparisons and rule-defined rewards for complex reasoning~\cite{mu2024rule, xie2025logicrl, wang2025kalman, xiong2025minimalist}. These structured RL principles are proving vital for training advanced reward models capable of nuanced multimodal understanding and for enabling RM self-improvement with systematic feedback~\cite{feng2025videor1, SPCT_2025}. Our \texttt{C-GRPO} builds on these trends, utilizing rule-based contrastive rewards from data augmentations to train a generative RM specific for EQA tasks.

\section{The \textsc{EQARewardBench} Dataset}
\label{sec:eqa_rm_bench}

To facilitate robust and standardized evaluation of reward models for Embodied Question Answering, we construct a new dataset \textsc{EQARewardBench}. This section details its generation pipeline, statistics, and splitting strategy.

\subsection{Dataset Generation Pipeline}
\label{sec:dataset_generation}

Our dataset construction process builds upon the OpenEQA~\cite{majumdar2024openeqa}, which provides instances comprising a question, ground truth answer, and associated episode memory. We extend this foundation in a two-step generation pipeline.

\noindent\textbf{Step 1: Diverse Response Generation.}
We first employ the Gemini-2.0-Flash~\cite{team2023gemini} model to generate a diverse set of responses for each EQA instance. Given the episode memory ($v^o$) and the original question ($q$) as input, this model produces multiple pairs of predicted answers ($a$) and accompanying reasoning traces ($z^o$). To ensure a rich dataset for subsequent reward modeling, we intentionally solicit a spectrum of predicted answers, encompassing both correct and incorrect responses, thereby fostering diversity.

\noindent\textbf{Step 2: High-Quality Score Generation.}
Next, Gemini-2.5-Pro-Experiment-03-14 generates detailed evaluations. For each Step 1 output tuple of \{episode memory ($v^o$), question ($q$), predicted answer ($a$), reasoning trace ($z^o$)\}, this model outputs a textual critique ($c_r$) and a scalar quality score. This score is an integer ranging from 0 to 10, indicating the quality of the predicted answer and reasoning trace. The ground truth answer input at this stage ensures accurate scores. After human verification for reliability, these scores become our ground truth scores ($s^{gt}$). The accompanying critiques ($c_r$), potentially influenced by this privileged input, are primarily for analysis and contextualizing $s_{gt}$, not for directly training reward models that must operate without such information during inference. This process yields a foundational dataset where each instance is a tuple: $\{q, a, v^o, z^o, c_r, s^{gt}\}$, forming the basis for both \textsc{EQARewardBench} and the \texttt{EQA-RM} fine-tuning data.

\subsection{Dataset Statistics and Splits}
\label{sec:dataset_stats_splits}

The episode memories ($v_o$) in our foundational dataset originate from two distinct 3D indoor environment collections provided by OpenEQA: HM3D~\cite{ramakrishnan2021hm3d} and ScanNet~\cite{dai2017scannet}. These sources contribute 697 and 1,546 instances in our dataset, respectively. 

To ensure robust model training, fair evaluation, and prevention of test set leakage, we partition this foundational dataset: \ding{182} \textbf{\textsc{EQARewardBench} ($D_B$)}: designed for evaluating diverse reward models, $D_B$ comprises all 823 HM3D instances and 713 ScanNet instances. \ding{183} \textbf{Fine-tuning Dataset ($D_F$)}: the remaining 697 distinct ScanNet instances are exclusively reserved for training our \texttt{EQA-RM} model.

Crucially, $D_F$ and $D_B$ maintain no overlap in underlying episode memory data (e.g., distinct scenes or trajectories), guaranteeing evaluation integrity. This splitting strategy enables comprehensive assessment: models trained on $D_F$ (ScanNet) are evaluated against the disjoint ScanNet portion of $D_B$ for in-distribution (ID) performance, and against its HM3D portion for out-of-distribution (OOD) generalization. Further details on dataset composition are provided in Appendix A.


\section{Methods}

This section details the methodology for training \texttt{EQA-RM} for nuanced evaluation of Embodied Question Answering trajectories. Our approach comprises two main stages: Rejective Fine-Tuning (RFT) to establish baseline capabilities in generating structured critiques and scores; followed by our novel Contrastive Group Relative Policy Optimization (\texttt{C-GRPO}) strategy to instill deeper sensitivities to critical aspects of trajectory quality. 

\subsection{Preliminaries and Notations}

We denote the generative reward model we aim to train as $R_\phi$. An instance consists of an input question $q$, a predicted answer $a$, an original reasoning trace $z^o=\{z^o_{1}, z^o_{2}, \dots, z^o_{T}\}$, and the original episode memory content $v^o$. The episode memory $v^o$ comprises a sequence of $N$ video frames, $v^o = \{v^o_{1}, v^o_{2}, \dots, v^o_{N}\}$. The desired output of $R_\phi$ for a given instance is an evaluation composed of a textual critique $c_r$ and a scalar score $s_r$. For each evaluated EQA instance, there is an associated ground truth score, denoted $s^{gt}$.

\subsection{Rejective Fine-Tuning}
\label{sec:rft}

After obtaining the foundational dataset (Section~\ref{sec:dataset_generation}), which includes ground truth scores ($s^{gt}$) for each EQA instance $\{q_i, a_i, z^o_{i}, v^o_{i}\}$, we initiate the training of \texttt{EQA-RM} ($R_\phi$) with Rejective Fine-Tuning (RFT). This first stage primarily aims to teach $R_\phi$ to generate outputs (a textual critique $c_r$ and a scalar score $s_r$) that conform to our desired structured format and exhibit a baseline level of quality. The RFT process involves two key steps:

\noindent\textbf{Step 1: Rejective Filtering.}
To construct the SFT dataset $D_{RFT}$, for each instance $\{q_i, a_i, z^o_{i}, v^o_{i}\}$ from finetuning dataset $D_F$, an auxiliary LLM evaluator $R^{aux}$ generates $N_{RFT}$ candidate evaluations. Each candidate evaluation consists of a critique $c^{aux}_{i,k}$ and a score $s^{aux}_{i,k}$. Importantly, $\{c^{aux}_{i,k}, s^{aux}_{i,k}\}$ are produced based only on the input $\{q_i, a_i, z^o_{i}, v^o_{i}\}$ without ground truth answer.
To ensure data quality for SFT, these candidate evaluations are rejective filtered. 

This rejective filtering process aims to remove "too easy" and "incorrect" candidate evaluations. 

An evaluation is considered correct ($\epsilon_{i,k} = 1$) if $|s^{gt}_{i} - s^{aux}_{i,k}| < \tau$ ($\tau$ is score tolerance), and $\epsilon_{i,k} = 0$ otherwise. Let $E_i = \sum_{k=1}^{N_{RFT}} \epsilon_{i,k}$ be the count of the correct evaluation of instance $i$. We select evaluations if:
\begin{equation}
\label{eq:rft_filter}\small
\epsilon_{i,k} \land \left( \neg (E_i = N_{RFT} \land N_{RFT} > 0) \right)
\end{equation}
This clause ensures an evaluation is included in $D_{RFT}$ only if it is "correct" (per $\epsilon_{i,k}=1$) and not all $N_{RFT}$ evaluations for instance $i$ were "correct" ("too easy"). Finally, $D_{RFT}$ comprising $\{q_i, a_i, z^o_{i}, v^o_{i}\}$ paired with critique and score $\{c_{i}, s_{i}\}$ that satisfy this rejective filtering process, is subsequently used for the initial SFT of $R_\phi$.

\noindent\textbf{Step 2: Supervised Fine-Tuning.}
The curated dataset $D_{RFT}$ is employed for SFT of our reward model \texttt{EQA-RM} $R_\phi$. For each sample in $D_{RFT}$, an input EQA instance comprising $(q_i, a_i, z^o_{i}, v^o_{i})$ is used to construct a multimodal prompt $P_i$, which also incorporates task-specific instructions. The corresponding output $T_i$ for $R_\phi$ is the structured string representation of the selected critique $c_i$ and score $s_i$ from $D_{RFT}$.
The SFT objective is to train $R_\phi$ by minimizing the negative log-likelihood loss $\mathcal{L}_{SFT}(\phi)$:
\begin{equation}
\label{eq:sft_loss}\small
\mathcal{L}_{SFT}(\phi) \! = \! - \!\!\!\! \sum_{\substack{(P_i, T_i) \\ \in D_{RFT} }} \! \sum_{t=1}^{|T_i|} \log P(T_{i,t} | P_i, T_{i,<t}; \phi)
\end{equation}
where $T_{i,t}$ is the $t$-th token of the target sequence $T_i$, and $T_{i,<t}$ represents the sequence of preceding tokens. This SFT stage primarily teaches $R_\phi$ to generate outputs in the specified format and establishes its baseline evaluative capability.

\begin{figure*}[htbp]
    \centering

    \begin{subfigure}[b]{0.48\textwidth}
        \centering
        \includegraphics[width=\linewidth]{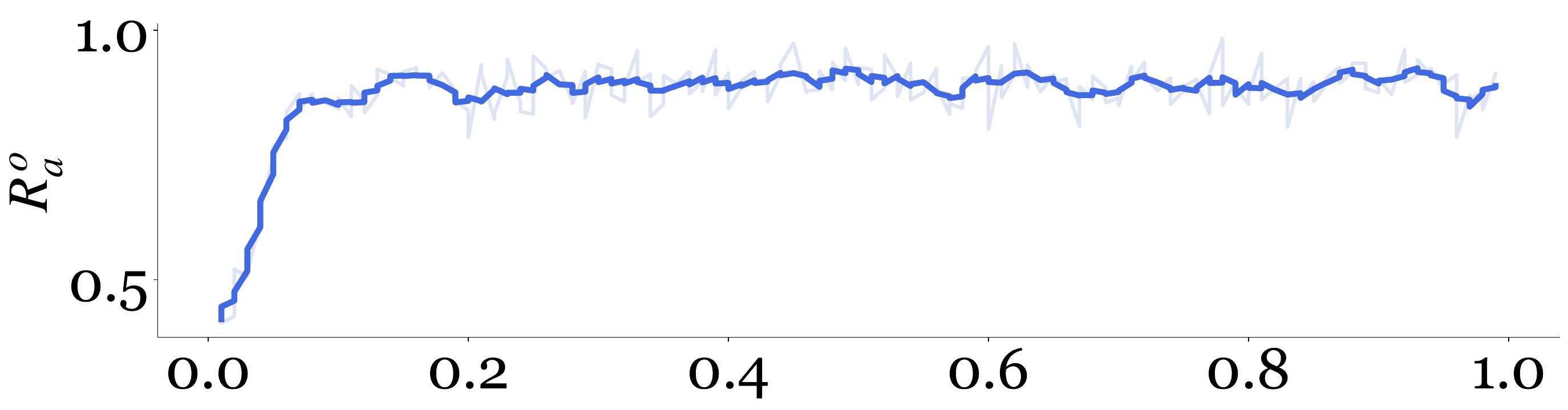}
        \vspace{-15pt}
        \caption{Accuracy Reward}
        \label{fig:accuracy_reward}
    \end{subfigure}
    \hfill
    \begin{subfigure}[b]{0.48\textwidth}
        \centering
        \includegraphics[width=\linewidth]{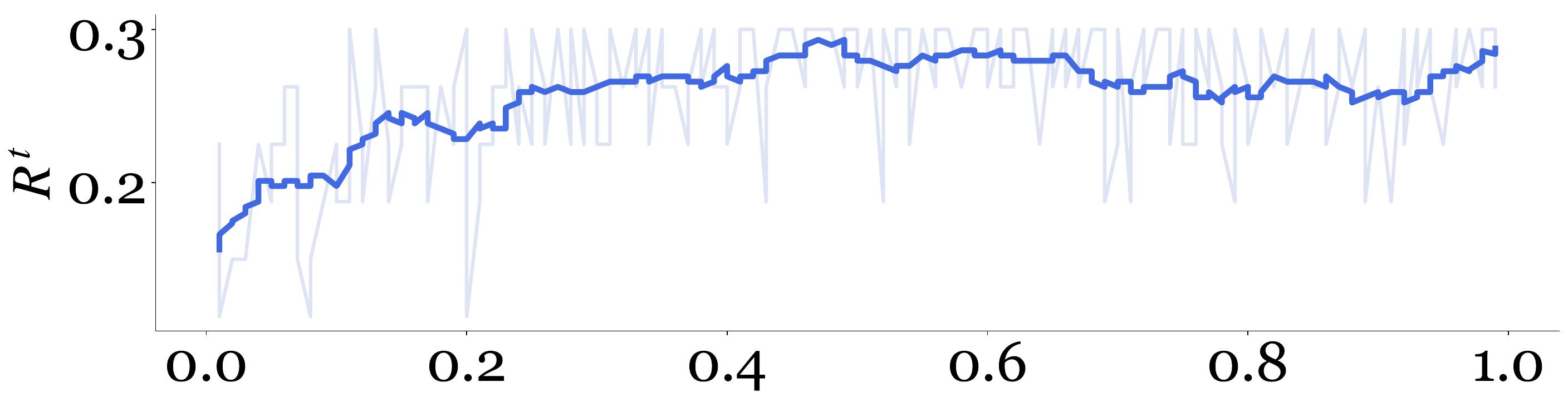}
        \vspace{-15pt}
        \caption{Reasoning Reward}
        \label{fig:reasoning_reward}
    \end{subfigure}


    \begin{subfigure}[b]{0.48\textwidth}
        \centering
        \includegraphics[width=\linewidth]{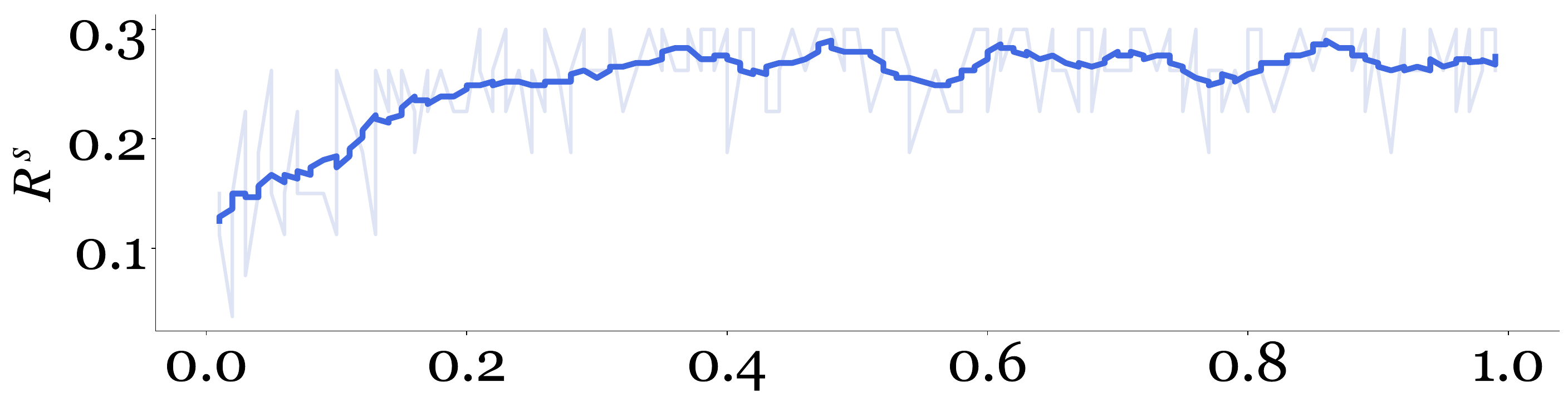}
        \vspace{-15pt}
        \caption{Spatial Reward}
        \label{fig:spatial_reward}
    \end{subfigure}
    \hfill
    \begin{subfigure}[b]{0.48\textwidth}
        \centering
        \includegraphics[width=\linewidth]{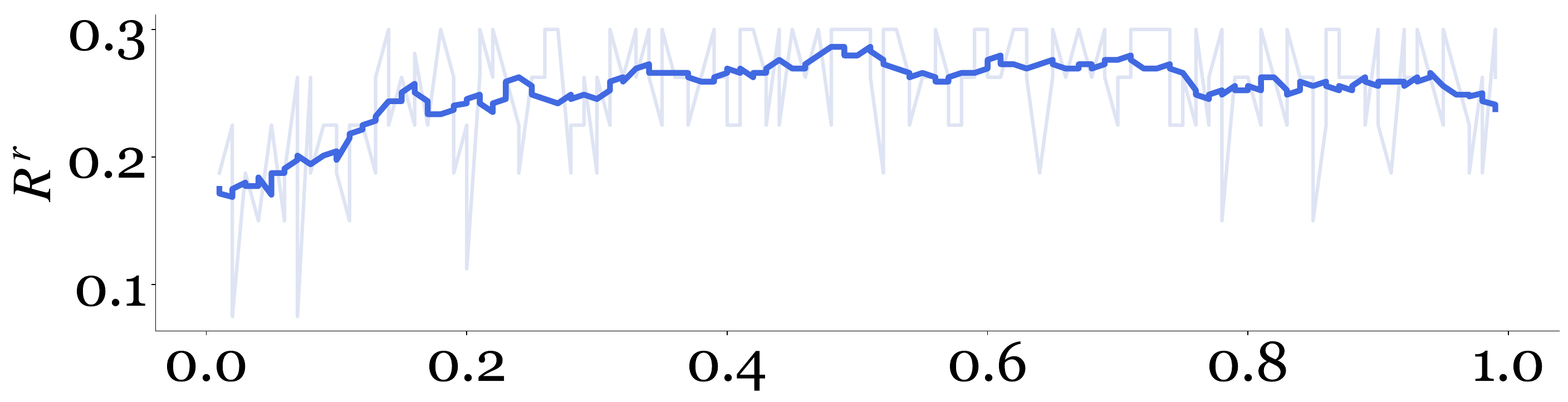}
        \vspace{-15pt}
        \caption{Temporal Reward}
        \label{fig:temporal_reward}
    \end{subfigure}
    \vspace{-3mm}
    \caption{Overview of training dynamics for \texttt{EQA-RM}'s core reward components of \texttt{C-GRPO}.}
    \label{fig:all_reward_trends}
\end{figure*}

\vspace{-1mm}
\subsection{Contrastive Group Relative Policy Optimization}
\label{sec:cgrpo}

While SFT establishes \texttt{EQA-RM}'s basic output formatting (critique $c_r$, score $s_r$), it often fails to instill deep comprehension of episode memory and reasoning from the limited finetuning data $D_F$. Our \textbf{Contrastive Group Relative Policy Optimization (\texttt{C-GRPO})} framework addresses this by using targeted contrastive rewards to explicitly train $R_\phi$ for crucial sensitivities: temporal visual ordering, fine-grained spatial details, and logical coherence of reasoning.

\texttt{C-GRPO} trains $R_\phi$ to differentiate its evaluation scores based on structured contrasts between original EQA instance components ($q_i, a_i, z^o_{i}, v^o_{i}$) and their synthetically perturbed counterparts. These perturbations include temporally shuffled video frames ($v^t_{i}$ from $v^o_{i}$), spatially masked frames ($v^s_{i}$ from $v^o_{i}$), and altered reasoning traces ($z^r_{i}$ from $z^o_{i}$). $R_\phi$ produces scores for the original condition ($S^o_{i}$) and for each corresponding augmented condition ($S^t_{i}, S^s_{i}, S^r_{i}$). For $S^t_{i}$ and $S^s_{i}$, $R_\phi$ scores the original $(q_i, a_i, z^o_{i})$ but is prompted as if the visual context were $v^t_{i}$ or $v^s_{i}$ respectively. For $S^r_{i}$, $R_\phi$ scores $(q_i, a_i)$ paired with the perturbed reasoning trace $z^r_{i}$ while using the original visual context $v^o_{i}$.

\noindent\textbf{Base Outcome Rewards.}
each evaluation score $s_{r,i}$ are assessed by the \textbf{Accuracy Reward} ($R_{a,i}$):
\begin{equation}\small
\label{eq:base_acc_reward}
R_{a,i}(s_{r,i}, s_{gt,i}) = \max(0, 1 - (\frac{|s_{r,k} - s_{gt,i}|}{10})^2)
\end{equation}
and \textbf{Format Reward} ($R_{f}$) is $1$ if text output adheres to the specified critique-score structure, and $0$ otherwise.

\noindent\textbf{\texttt{C-GRPO} Contrastive Rewards.}
For an $x$-augmented version ($x \in \{t,s,r\}$ for temporal, spatial, reasoning respectively), let $R^x_{a,i}$ be the corresponding accuracy reward.
This mechanism conditionally boosts the $R^o_{a,i}$ for evaluations of original instances. For each active augmentation type $x \in \{t,s,r\}$:
Let $\overline{R^o_{a}}$ and $\overline{R^x_{a}}$ be the batch-mean accuracy rewards (Eq.~\ref{eq:base_acc_reward}) for original and $x$-augmented evaluations, respectively. 
The per-evaluation boost $R^x$ for the $i$-th original evaluation is then defined as:
\begin{equation}\small
\label{eq:contrastive_boost_bxk}
R^x_{i} = \begin{cases} \mu, & \text{if } \overline{R^o_{a}} \ge \delta \cdot \overline{R^x_{a}} \\ 0, & \text{otherwise} \end{cases}
\end{equation}
where $\delta=0.95$ and $\mu=0.3$ are hyperparameters. This yields boosts $R^t, R^s, R^r$ for active contrasts.

\noindent\textbf{Total Reward.}
The total reward is:
\begin{equation}\small
\label{eq:cgrpo_total_reward}
R^A_i = R^o_{a} + R^o_{f} + (R^t+R^s+R^r)/3
\end{equation}

\noindent\textbf{Advantage and Policy Update.}
The advantage $A_i$ for the $i$-th evaluation (from $G$ total evaluations) is computed by normalizing the total rewards $\{R^A_j\}_{j=1}^G$ within the group:
\begin{equation}\small
A_i = \frac{R_i^A - \mathrm{mean}(\{R_j^A\}_{j=1}^G)}{\mathrm{std}(\{R_j^A\}_{j=1}^G) + \epsilon}
\label{eq:cgrpo_advantage}
\end{equation}
where $\epsilon$ is a small constant for numerical stability. $R_\phi$ are updated follows the objective of \texttt{C-GRPO}:
\begin{equation}\small
\label{eq:cgrpo_loss}
\begin{split}
& \mathcal{J}_{\text{\texttt{C-GRPO}}}(\phi) = 
\mathbb{E}_{d_i \sim D_F} \Bigg[ \frac{1}{G} \sum_{k=1}^G \bigg( \\
& \min \Big( r_k(\phi) A_k, \text{clip}( r_k(\phi), 1 - \epsilon_c, 1 + \epsilon_c ) A_k \Big) \\
& - \beta_K \, \mathbb{D}_{\mathrm{KL}}(P_\phi \| P_{\phi_{\mathrm{ref}}}) \bigg) \Bigg]
\end{split}
\end{equation}
where $r_k(\phi)$ is the probability ratio, $\phi_{\text{ref}}$ are parameters before the update, $\epsilon_c$ is a clipping hyperparameter, and $R_{\phi_{\mathrm{ref}}}$ is a reference model (typically the SFT version of $R_\phi$).

\section{Experiments}
\label{sec:experiments}

In this section, our goal is to assess the effectiveness of \texttt{EQA-RM} and answer the following questions:

\begin{itemize}[topsep=0pt, leftmargin=10pt, itemsep=-2pt, partopsep=0pt, parsep=0pt]
    \item \textbf{RQ1}: How does \texttt{EQA-RM} perform compared to existing Visual-based Reward Models?
    \item \textbf{RQ2}: How does the performance of \texttt{EQA-RM} scale with increased test-time compute?
    \item \textbf{RQ3}: What is the impact of each component of the proposed reward strategy on \texttt{EQARewardBench}'s performance?
\end{itemize}

\begin{table*}[t] 
\centering 
\caption{Main evaluation results on the EQARewardBench dataset. We report Accuracy (Acc) and RMSE for each method across two environments (HM3D and ScanNet), and their aggregated Overall performance. All results are shown with differences relative to Qwen2-VL-2B-Instruct model (Better is marked in red; worse is marked in blue).} 
\label{tab:main_eqa_rm_results} 
\resizebox{\textwidth}{!}{ 
\begin{NiceTabular}{@{}lcccccc@{}}[code-before = \rectanglecolor{headergray}{1-1}{2-7} \rectanglecolor{rowgray}{4-1}{4-7} \rectanglecolor{rowgray}{6-1}{6-7} \rectanglecolor{rowgray}{9-1}{9-7} \rectanglecolor{rowgray}{12-1}{12-7} \rectanglecolor{eqagray}{13-1}{13-7}]
\toprule 
\multirow{2}{*}{\textbf{Models}} & \multicolumn{2}{c}{\textbf{EQARewardBench-HM3D}} & \multicolumn{2}{c}{\textbf{EQARewardBench-ScanNet}} & \multicolumn{2}{c}{\textbf{Overall}} \\ 
\cmidrule(lr){2-3} \cmidrule(lr){4-5} \cmidrule(lr){6-7} 
& \textbf{Acc ↑} & \textbf{RMSE ↓} & \textbf{Acc ↑} & \textbf{RMSE ↓} & \textbf{Acc ↑} & \textbf{RMSE ↓} \\ 
\midrule 
\multicolumn{7}{c}{\textit{VLM-as-a-Judge}} \\ 
\midrule 
Qwen2-VL-2B-Instruct & 33.71 & 3.836 & 32.44 & 4.158 & 33.08 & 3.997 \\ 
Gemini-2.5-Flash & \textbf{61.25}\accBetter{27.54} & 4.872\rmseWorse{1.036} & \underline{58.33}\accBetter{25.89} & 3.805\rmseBetter{0.353} & \underline{59.79}\accBetter{26.71} & 4.339\rmseWorse{0.342} \\ 
GPT-4o & \underline{60.17}\accBetter{26.46} & 6.847\rmseWorse{3.011} & 56.72\accBetter{24.28} & 5.313\rmseWorse{1.155} & 58.45\accBetter{25.37} & 6.080\rmseWorse{2.083} \\ 
Claude-3.5-Haiku & 57.88\accBetter{24.17} & 5.316\rmseWorse{1.480} & 54.12\accBetter{21.68} & 4.291\rmseWorse{0.133} & 56.00\accBetter{22.92} & 4.804\rmseWorse{0.807} \\ 
\midrule 
\multicolumn{7}{c}{\textit{Standard Visual-Based Reward Models}} \\ 
\midrule 
RoVRM & 38.12\accBetter{4.41} & 3.341\rmseBetter{0.495} & 40.23\accBetter{7.79} & 3.158\rmseBetter{1.000} & 39.18\accBetter{6.10} & 3.250\rmseBetter{0.747} \\ 
VisualPRM & 40.07\accBetter{6.36} & 3.562\rmseBetter{0.274} & 37.41\accBetter{4.97} & 3.413\rmseBetter{0.745} & 38.74\accBetter{5.66} & 3.488\rmseBetter{0.509} \\ 
\midrule 
\multicolumn{7}{c}{\textit{Proposed Method}} \\ 
\midrule 
base \texttt{EQA-RM} & 39.16\accBetter{5.45} & \underline{3.081}\rmseBetter{0.755} & 45.78\accBetter{13.34} & \underline{2.826}\rmseBetter{1.332} & 42.47\accBetter{9.39} & \underline{2.953}\rmseBetter{1.044} \\ 
\quad w/ test scaling (K=32) & 58.65\accBetter{24.94} & \textbf{2.918}\rmseBetter{0.918} & \textbf{65.06}\accBetter{32.62} & \textbf{2.537}\rmseBetter{1.621} & \textbf{61.86}\accBetter{28.78} & \textbf{2.727}\rmseBetter{1.270} \\ 
\bottomrule 
\end{NiceTabular}%
} 
\vspace{-3mm} 
\end{table*}

\subsection{Experiment Setup}

\noindent\textbf{Dataset and Benchmark.} 
All experimental evaluations are conducted using our \textsc{EQARewardBench} ($D_B$). The methodology behind it is detailed in Section~\ref{sec:eqa_rm_bench}. The evaluation focuses on the Verifier mode, assessing the RM's capability to accurately score pre-generated policy responses based on the offline-annotated ground truth scores.

\noindent\textbf{Base Model.}
Our base model is Qwen2-VL-2B-Instruct~\cite{wang2024qwen2}, instruction-tuned for robust multimodal understanding.

\noindent\textbf{Baselines.}
We compare \texttt{EQA-RM} against a comprehensive set of baseline methods to establish its relative effectiveness:
\textbf{(1) VLM-as-a-Judge:} State-of-the-art VLMs prompted offline to evaluate EQA responses on our test set. Including Gemini-2.5-Flash~\cite{team2023gemini}, GPT-4o~\cite{hurst2024gpt} and Claude-3.5-Haiku. 
\textbf{(2) Generic Standard RMs:} Adapt existing VLMs or Visual RMs including RoVRM~\cite{wang2025rovrm}, VisualPRM~\cite{wang2025visualprm}.

\begin{table*}[t]
\centering
\caption{Accuracy performance on the EQARewardBench-Scannet dataset, broken down by EQA question type. Higher is better for all metrics. All results are shown with differences relative to the Qwen2-VL-2B-Instruct model (Better is marked in red; Worse is marked in blue).}
\label{tab:question_type_acc_performance}
\vspace{-1mm}
\resizebox{\textwidth}{!}{
\begin{NiceTabular}{@{}l*{7}{c}@{}}[code-before = 
\rectanglecolor{headergray}{1-1}{2-8}
\rectanglecolor{rowgray}{5-1}{5-8}
\rectanglecolor{rowgray}{7-1}{7-8}
\rectanglecolor{rowgray}{9-1}{9-8}
\rectanglecolor{rowgray}{12-1}{12-8}
\rectanglecolor{eqagray}{13-1}{13-8}
]
\toprule
\multirow{2}{*}{\textbf{Models}} & \Block[c]{2-1}{\textbf{Object}\\\textbf{Recognition}} & \Block[c]{2-1}{\textbf{Object}\\\textbf{Localization}} & \Block[c]{2-1}{\textbf{Attribute}\\\textbf{Recognition}} & \Block[c]{2-1}{\textbf{Spatial}\\\textbf{Understanding}} & \Block[c]{2-1}{\textbf{Object State}\\\textbf{Recognition}} & \Block[c]{2-1}{\textbf{Functional}\\\textbf{Reasoning}} & \Block[c]{2-1}{\textbf{World}\\\textbf{Knowledge}} \\
&&&&&&& \\
\midrule
\multicolumn{8}{c}{\textit{VLM-as-a-Judge}} \\
\midrule
Qwen2-VL-2B-Instruct & 30.30 & 36.76 & 38.69 & 29.41 & 29.88 & 38.71 & 27.34 \\
Gemini-2.5-Flash & \textbf{69.23}\accBetter{38.93} & \underline{62.15}\accBetter{25.39} & 61.07\accBetter{22.38} & 38.92\accBetter{9.51} & \textbf{51.88}\accBetter{22.00} & \underline{64.05}\accBetter{25.34} & \textbf{61.51}\accBetter{34.17} \\
GPT-4o & 66.39\accBetter{36.09} & 60.00\accBetter{23.24} & 59.84\accBetter{21.15} & \textbf{41.51}\accBetter{12.10} & \underline{50.00}\accBetter{20.12} & 57.39\accBetter{18.68} & \underline{60.68}\accBetter{33.34} \\
Claude-3.5-Haiku & 60.15\accBetter{29.85} & 56.76\accBetter{20.00} & \underline{64.22}\accBetter{25.53} & 35.83\accBetter{6.42} & 43.51\accBetter{13.63} & 53.88\accBetter{15.17} & 60.11\accBetter{32.77} \\
\midrule
\multicolumn{8}{c}{\textit{Standard Visual-Based Reward Models}} \\
\midrule
RoVRM & 48.90\accBetter{18.60} & 50.13\accBetter{13.37} & 47.82\accBetter{9.13} & 25.91\accWorse{3.50} & 36.50\accBetter{6.62} & 44.69\accBetter{5.98} & 27.38\accBetter{0.04} \\
VisualPRM & 29.60\accWorse{1.30} & 42.90\accBetter{6.14} & 43.62\accBetter{4.93} & \underline{39.21}\accBetter{9.80} & 33.93\accBetter{4.05} & 41.33\accBetter{2.62} & 29.34\accBetter{2.00} \\
\midrule
\multicolumn{8}{c}{\textit{Proposed Method}} \\
\midrule
\texttt{base \texttt{EQA-RM}} & 36.36\accBetter{6.06} & 50.81\accBetter{14.05} & 51.82\accBetter{13.13} & 26.05\accWorse{3.36} & 21.95\accWorse{7.93} & 45.97\accBetter{7.26} & 35.94\accBetter{8.60} \\
\quad w/ test scaling (K=32) & \underline{68.18}\accBetter{37.88} & \textbf{69.73}\accBetter{32.97} & \textbf{71.53}\accBetter{32.84} & 37.82\accBetter{8.41} & 37.20\accBetter{7.32} & \textbf{66.94}\accBetter{28.23} & 57.81\accBetter{30.47} \\
\bottomrule
\end{NiceTabular}%
}
\vspace{-1mm}
\end{table*}

\noindent\textbf{Evaluation Metrics.}
We evaluate the performance of all reward models using the following metrics. 
\begin{itemize}[topsep=0pt, leftmargin=10pt, itemsep=-2pt, partopsep=0pt, parsep=0pt]
    \item \textbf{Accuracy} measures the proportion of predictions where the gap between the predicted score $s^p$ and the target score $s^{gt}$ is within a predefined tolerance $\tau$. We use the tolerance $\tau=2$ in the experiments:
    
    \begin{equation}\small
    \setlength\abovedisplayskip{25pt} 
        \text{Acc} = \frac{1}{N} \sum_{i=1}^{N} \mathbb{I}(|s^p_{i} - s^{gt}_{i}| \le \tau)
    \end{equation}
    \item \textbf{Root Mean Square Error} quantifies the average magnitude of the error between predicted and target scores:
    \begin{equation}\small
        \text{RMSE} = \sqrt{\frac{1}{N} \sum_{i=1}^{N} (s^p_{i} - s^{gt}_{i})^2}
    \end{equation}

\end{itemize}

\noindent\textbf{Implementation Details}.
Training utilized the AdamW optimizer with a learning rate of $1e-6$ and a batch size of 1. 
Keyframe selection during training and test to select $N=5$ frames for each episode history video. 
Further details are provided in the Appendix B.

\noindent\textbf{Test-time Scaling Configuration}.
To enhance \texttt{EQA-RM}'s evaluation robustness, our test-time scaling (TTS) strategy involves sampling K diverse evaluative reasoning paths from a given EQA trajectory. This is facilitated by a temperature setting of 0.8 and top-p sampling with $p=0.9$. We explore K values of 1 (no TTS), 2, 4, 8, 16, and 32. These K assessments are aggregated using either Majority Voting or Averaging Rewards for continuous scores. This TTS approach enables \texttt{EQA-RM} to synthesize multiple perspectives, aiming for a more reliable and comprehensive trajectory assessment by mitigating single-inference pass limitations.

\subsection{Main Results}
\label{subsec:main_results}
The overall performance of various reward models on the EQARewardBench dataset is presented in Table~\ref{tab:main_eqa_rm_results}. Our proposed \texttt{EQA-RM}, fine-tuned from Qwen2-VL-2B-Instruct, is benchmarked against its base model, other VLM-as-a-Judge methods, and Standard Visual-Based Reward Models.
\texttt{EQA-RM} demonstrates substantial gains over its base Qwen2-VL-2B-Instruct model, for instance, improving overall accuracy by over 9\% while also achieving a markedly lower RMSE. It also consistently outperforms standard visual-based RMs like RoVRM and VisualPRM in both overall accuracy and RMSE, showing the advantage of our specialized training.

While the base \texttt{EQA-RM}'s accuracy is lower than leading VLM-as-a-Judge models such as GPT-4o and Gemini-2.5-Flash, it provides more precise reward signals, evidenced by its considerably lower RMSE. Crucially, with test-time scaling (K=32), \texttt{EQA-RM}'s accuracy significantly increases to an overall score of approximately 61.9\%, surpassing these top VLM judges and achieving the best RMSE. This positions the scaled \texttt{EQA-RM} as the top-performing model on our benchmark.

\subsection{Test-time Scalability}
\label{subsec:tts_results}

\begin{figure*}[t!]
    \centering

    \newlength{\subfigindividualwidth}
    \setlength{\subfigindividualwidth}{0.25\textwidth} 
    \newlength{\subfigcontenttotalwidth}
    \setlength{\subfigcontenttotalwidth}{4\subfigindividualwidth}

    \newlength{\subfiginterspace}
    \setlength{\subfiginterspace}{(\textwidth - \subfigcontenttotalwidth) / 3}

    \begin{subfigure}[b]{\subfigindividualwidth}
        \centering
        \includegraphics[width=\linewidth]{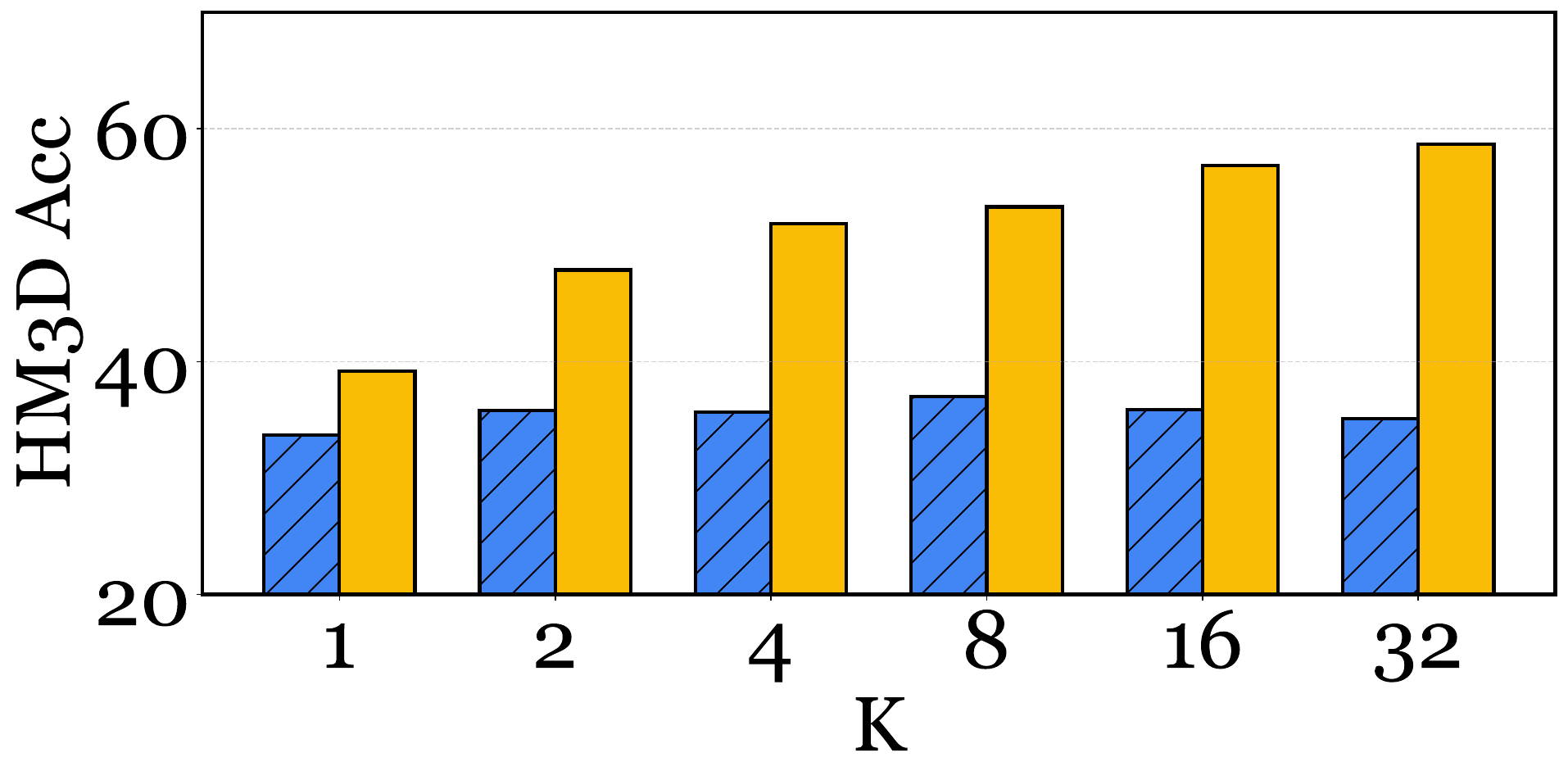}
        \label{fig:hm3d_acc_bar}
    \end{subfigure}%
    \hspace{\subfiginterspace}%
    \begin{subfigure}[b]{\subfigindividualwidth}
        \centering
        \includegraphics[width=\linewidth]{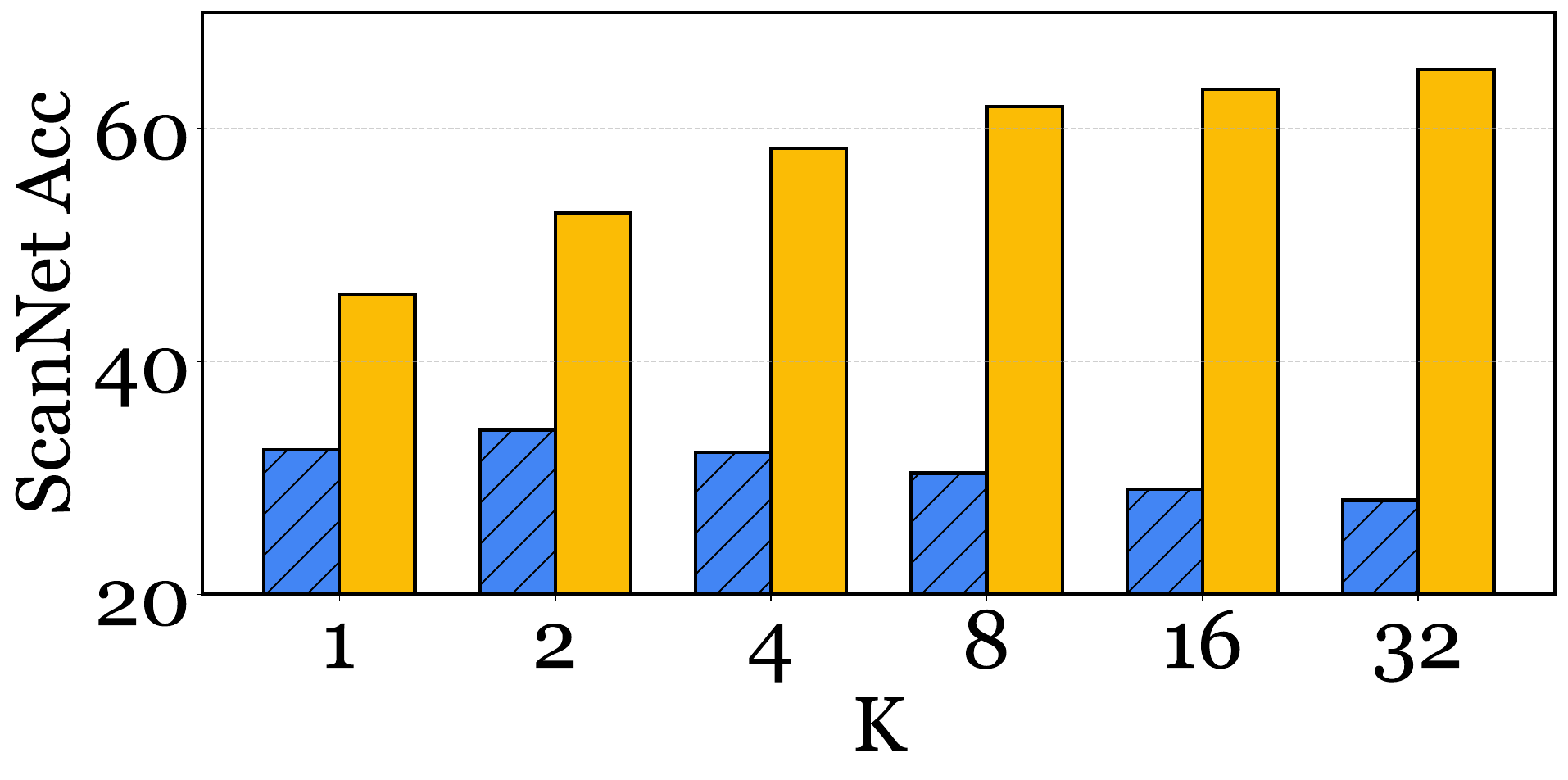}
        \label{fig:scannet_acc_bar}
    \end{subfigure}%
    \hspace{\subfiginterspace}%
    \begin{subfigure}[b]{\subfigindividualwidth}
        \centering
        \includegraphics[width=\linewidth]{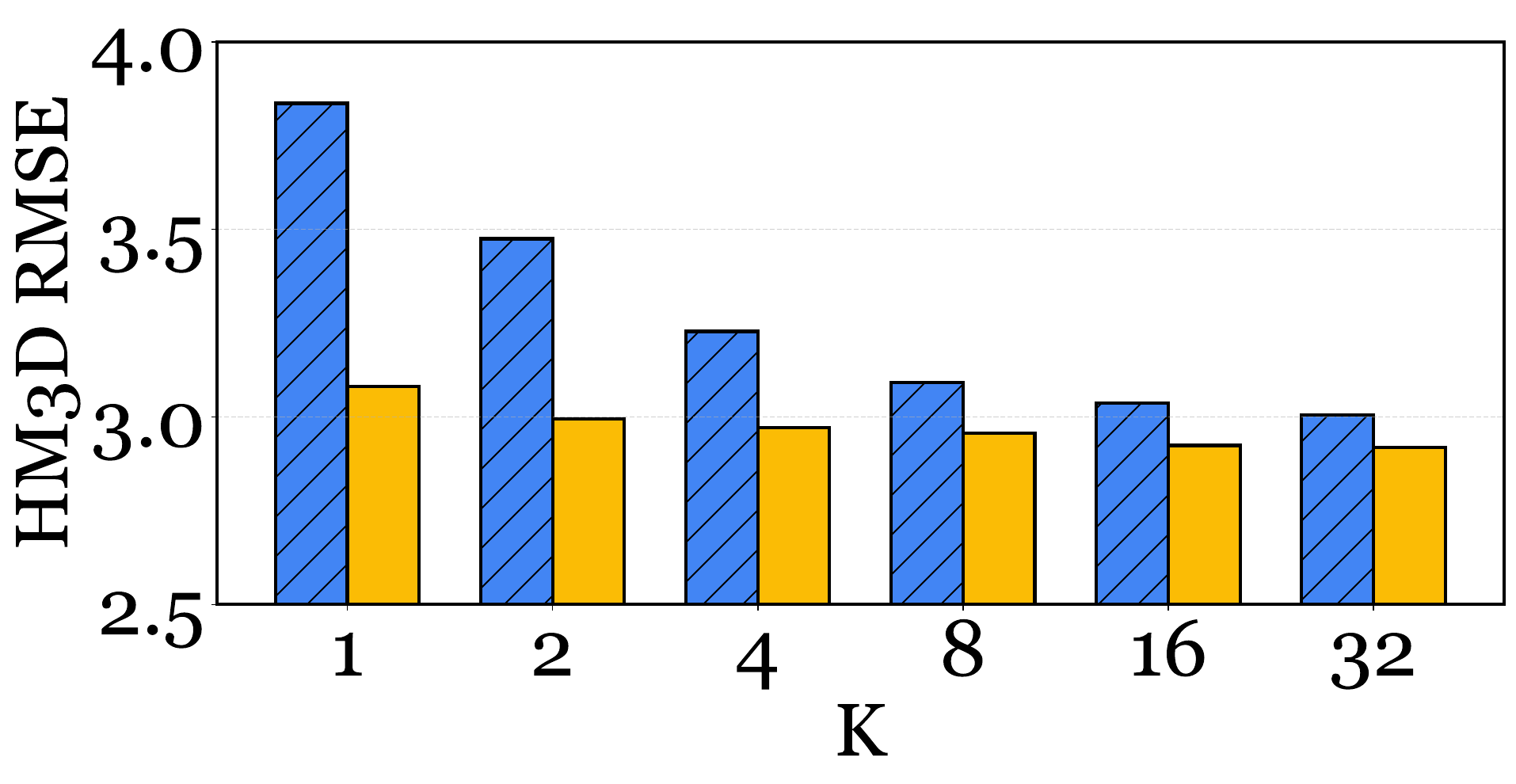}
        \label{fig:hm3d_rmse_bar}
    \end{subfigure}%
    \hspace{\subfiginterspace}%
    \begin{subfigure}[b]{\subfigindividualwidth}
        \centering
        \includegraphics[width=\linewidth]{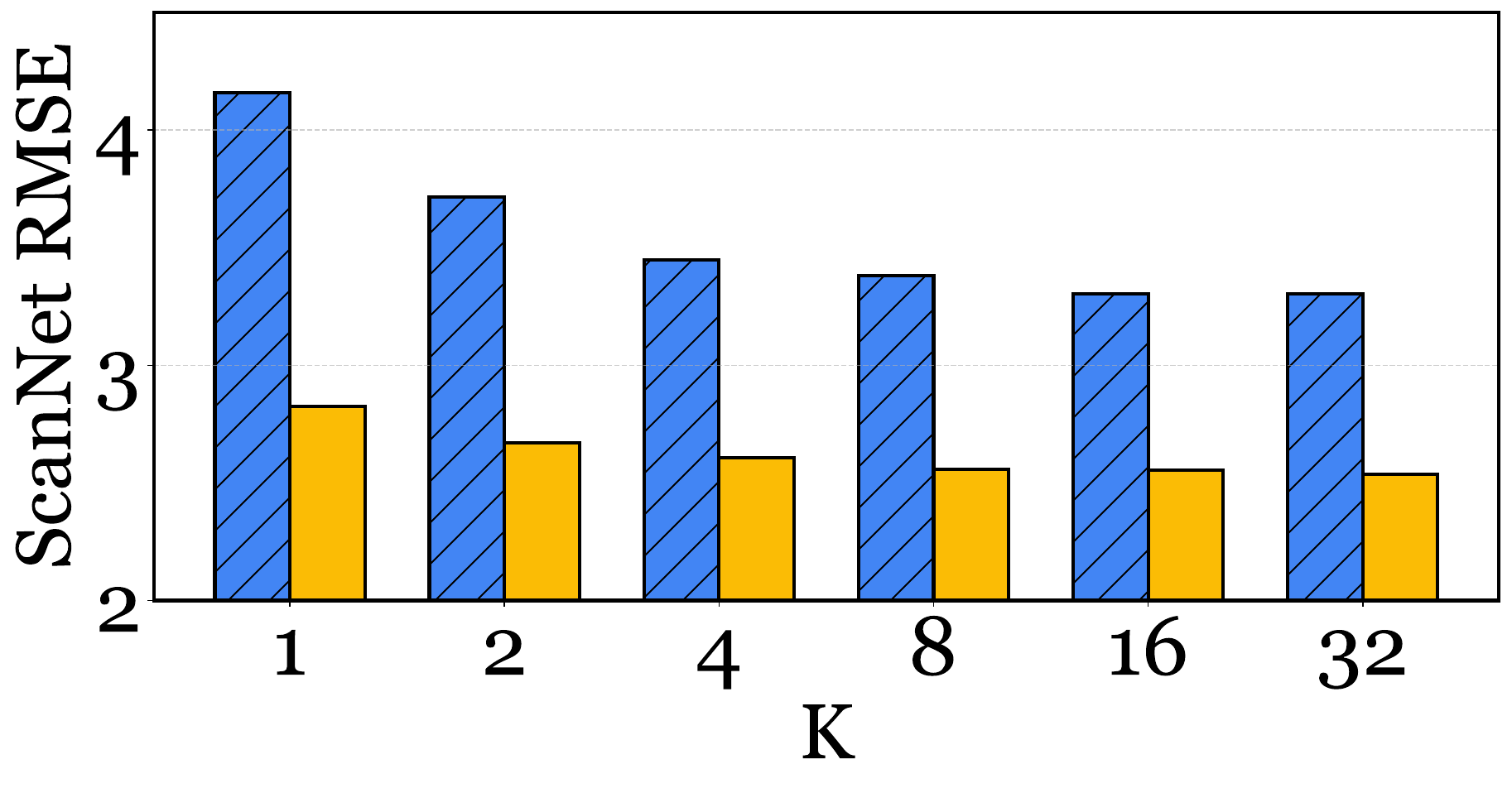}
        \label{fig:scannet_rmse_bar}
    \end{subfigure}
    \vspace{-35pt}
    \caption{Effect of test-time scaling (varying K) on \texttt{EQA-RM} performance, comparing \texttt{EQA-RM} and Qwen2-VL-2B-Instruct on HM3D and ScanNet datasets.}
    \label{fig:tts}
\end{figure*}

The benefits of test-time scaling for \texttt{EQA-RM} are demonstrated in Figure~\ref{fig:tts}. As the number of sampled critiques (K) increases from 1 to 32, \texttt{EQA-RM} exhibits a strong and consistent improvement in performance. Overall accuracy sees substantial gains. For example, increasing by over 19\% on both ScanNet and HM3D datasets, alongside a steady decrease in RMSE, indicating enhanced prediction quality. This robust scalability contrasts sharply with the base Qwen2-VL-2B-Instruct model. Although the latter shows test-time scaling on RMSE, But it does not exhibit a positive trend in terms of accuracy with increasing K. This highlights that the test-time scaling advantage is a specific outcome of our  training methodology for \texttt{EQA-RM}.

\begin{table*}[t!]
\centering
\caption{Ablation study on core components and reward formulations of \texttt{EQA-RM}. $R^b$ denotes the base accuracy and format reward. $R^t$, $R^s$, and $R^r$ represent the contrastive temporal, spatial and reasoning reward, respectively (defined in Section~\ref{sec:cgrpo}). Configurations in part B are subsequent to the RFT stage and detail the specific reward formula optimized by \texttt{C-GRPO}. All results are shown with differences relative to the Base Model.}
\label{tab:ablation_studies}
\resizebox{\textwidth}{!}{%
\begin{NiceTabular}{@{}lcccccc@{}}[code-before = 
\rectanglecolor{headergray}{1-1}{2-7}
\rectanglecolor{rowgray}{4-1}{4-7}
\rectanglecolor{rowgray}{7-1}{7-7}
\rectanglecolor{rowgray}{9-1}{9-7}
\rectanglecolor{rowgray}{11-1}{11-7}
\rectanglecolor{rowgray}{13-1}{13-7}
\rectanglecolor{rowgray}{15-1}{15-7}
\rectanglecolor{rowgray}{17-1}{17-7}
\rectanglecolor{rowgray}{19-1}{19-7}
\rectanglecolor{eqagray}{20-1}{20-7}
]
\toprule
\multirow{2}{*}{\textbf{Method Configuration}} & \multicolumn{2}{c}{\textbf{EQARewardBench-HM3D}} & \multicolumn{2}{c}{\textbf{EQARewardBench-ScanNet}} & \multicolumn{2}{c}{\textbf{Overall}} \\
\cmidrule(lr){2-3} \cmidrule(lr){4-5} \cmidrule(lr){6-7}
& Acc (\%) $\uparrow$ & RMSE $\downarrow$ & Acc (\%) $\uparrow$ & RMSE $\downarrow$ & Acc (\%) $\uparrow$ & RMSE $\downarrow$ \\
\midrule
\multicolumn{7}{@{}l}{\textit{A. Core Training Stage and Architectural Ablations:}} \\
Base Model (Qwen2-VL-2B-Instruct) & 33.71 & 3.836 & 32.44 & 4.158 & 33.08 & 3.997 \\
RFT only & 32.14\accWorse{1.57} & 3.781\rmseBetter{0.055} & 31.24\accWorse{1.20} & 3.879\rmseBetter{0.279} & 31.69\accWorse{1.39} & 3.830\rmseBetter{0.167} \\
\midrule
\multicolumn{7}{@{}l}{\textit{B. Specific Reward Formulations (without RFT):}} \\
RL ($R_b$ only) & 29.12\accWorse{4.59} & 4.132\rmseWorse{0.296} & 30.54\accWorse{1.90} & 4.035\rmseBetter{0.123} & 29.83\accWorse{3.25} & 4.084\rmseWorse{0.087} \\
RL ($R_b + R_t$) (+ Temporal) & 31.55\accWorse{2.16} & 4.027\rmseWorse{0.191} & 32.85\accBetter{0.41} & 3.953\rmseBetter{0.205} & 32.20\accWorse{0.88} & 3.990\rmseBetter{0.007} \\
RL ($R_b + R_s$) (+ Spatial) & 28.95\accWorse{4.76} & 3.759\rmseBetter{0.077} & 30.15\accWorse{2.29} & 3.788\rmseBetter{0.370} & 29.55\accWorse{3.53} & 3.774\rmseBetter{0.223} \\
RL ($R_b + R_r$) (+ Reasoning) & 32.05\accWorse{1.66} & 3.801\rmseBetter{0.035} & 33.57\accBetter{1.13} & 3.882\rmseBetter{0.276} & 32.81\accWorse{0.27} & 3.842\rmseBetter{0.155} \\
RL ($R_b + R_t + R_s + R_r$) & 31.24\accWorse{2.47} & 3.812\rmseBetter{0.024} & 32.94\accBetter{0.50} & 3.935\rmseBetter{0.223} & 32.09\accWorse{0.99} & 3.874\rmseBetter{0.123} \\
\midrule
\multicolumn{7}{@{}l}{\textit{C. Specific Reward Formulations (with RFT):}} \\
RFT + RL ($R_b$ only) & 33.50\accWorse{0.21} & 3.895\rmseWorse{0.059} & 31.80\accWorse{0.64} & 4.189\rmseWorse{0.031} & 32.65\accWorse{0.43} & 4.042\rmseWorse{0.045} \\
RFT + RL ($R_b + R_t$) (+ Temporal) & 32.95\accWorse{0.76} & 3.750\rmseBetter{0.086} & 35.20\accBetter{2.76} & 4.210\rmseWorse{0.052} & 34.08\accBetter{1.00} & 3.980\rmseBetter{0.017} \\
RFT + RL ($R_b + R_s$) (+ Spatial) & 30.15\accWorse{3.56} & 3.915\rmseWorse{0.079} & 29.50\accWorse{2.94} & 4.233\rmseWorse{0.075} & 29.83\accWorse{3.25} & 4.074\rmseWorse{0.077} \\
RFT + RL ($R_b + R_r$) (+ Reasoning) & \underline{39.30}\accBetter{5.59} & 3.152\rmseBetter{0.684} & 45.50\accBetter{13.06} & 2.984\rmseBetter{1.174} & 42.40\accBetter{9.32} & 3.068\rmseBetter{0.929} \\
RFT + RL ($R_b + R_t + R_s$) (+ Temporal + Spatial) & 38.27\accBetter{4.56} & \underline{3.015}\rmseBetter{0.821} & 42.76\accBetter{10.32} & \underline{2.755}\rmseBetter{1.403} & 40.52\accBetter{7.44} & \underline{2.885}\rmseBetter{1.112} \\
RFT + RL ($R_b + R_t + R_r$) (+ Temporal + Reasoning) & 39.10\accBetter{5.39} & 3.102\rmseBetter{0.734} & \textbf{45.80}\accBetter{13.36} & 2.907\rmseBetter{1.251} & \underline{42.45}\accBetter{9.37} & 3.005\rmseBetter{0.992} \\
RFT + RL ($R_b + R_s + R_r$) (+ Spatial + Reasoning) & \textbf{39.45}\accBetter{5.74} & \textbf{2.976}\rmseBetter{0.860} & 45.43\accBetter{12.99} & \textbf{2.714}\rmseBetter{1.444} & 42.44\accBetter{9.36} & \textbf{2.845}\rmseBetter{1.152} \\
\textbf{RFT + RL ($R_b + R_t + R_s + R_r$) (\texttt{EQA-RM})} & 39.16\accBetter{5.45} & 3.081\rmseBetter{0.755} & \underline{45.78}\accBetter{13.34} & 2.826\rmseBetter{1.332} & \textbf{42.47}\accBetter{9.39} & 2.954\rmseBetter{1.043} \\
\bottomrule
\end{NiceTabular}%
}
\vspace{-2mm}
\end{table*}

\subsection{Ablation Studies} 
\label{subsec:ablations}

Table~\ref{tab:ablation_studies} presents our ablation study on \textsc{EQARewardBench}, reporting accuracy and RMSE across its HM3D and ScanNet subsets.

\noindent\textbf{Training Stage Analysis.} Part A indicates that RFT alone slightly reduces overall accuracy (-1.39\%) while marginally improving RMSE compared to the base model. This is likely because RFT, with its limited samples, primarily focuses on teaching output formatting, which, if used in isolation, may not enhance broader understanding.

\noindent\textbf{Reward Components.} Ablating RFT pre-training (Part B) reveals that most subsequent C-GRPO reward formulations underperform the base model in accuracy, underscoring RFT's critical role for effective initialization. With RFT pre-training (Part C), the reasoning reward ($R_r$) alone yields the largest single-component accuracy increase (+9.32\%). Conversely, the spatial reward ($R_s$), while detrimental in isolation (-3.25\%), improves performance when combined with other components, suggesting a regularizing effect against overfitting to specific (e.g., temporal or reasoning) cues. These results affirm the efficacy of our contrastive reward design and the importance of RFT cold start for optimal performance.

\subsection{Question Type Performance Analysis}
\label{subsubsec:perf_by_q_type}
Table~\ref{tab:question_type_acc_performance} presents an accuracy breakdown by EQA question type on the \textsc{EQARewardBench-ScanNet} dataset. Our \texttt{EQA-RM} with test-time scaling (K=32) demonstrates notable strength, achieving top performance in Object Localization, Attribute Recognition, and Functional Reasoning. It also proves highly competitive in Object Recognition and World Knowledge against strong VLM-as-a-Judge baselines. This scaled version of \texttt{EQA-RM} significantly improves upon its base, surpassing several VLM judges in these key areas.
\section{Conclusion}

We introduced \texttt{EQA-RM}, a generative reward model tailored for nuanced evaluation of complex Embodied Question Answering (EQA) trajectories, alongside \textsc{EQARewardBench}, a dedicated benchmark for this task. Trained with our novel Contrastive Group Relative Policy Optimization (\texttt{C-GRPO}) strategy, \texttt{EQA-RM} learns to assess critical spatial, temporal, and reasoning understanding. Empirical results demonstrate \texttt{EQA-RM}'s effectiveness and high sample efficiency, achieving 61.84\% accuracy on \textsc{EQARewardBench} through test-time scaling, thereby outperforming strong proprietary and open-source baselines. This work presents a significant step towards robust reward modeling in embodied AI, offering tools and methodologies to foster more capable EQA agents.

\noindent\textbf{Acknowledgement.} This work is partially supported by Amazon Research Award, Cisco Faculty Award, UNC Accelerating AI Awards, NAIRR Pilot Award, OpenAI Researcher Access Award, and Gemma Academic Program GCP Credit Award.


\section*{Limitations and Future Work}
\label{sec:limitations_future_work}

While \texttt{EQA-RM} and our C-GRPO strategy demonstrate significant advancements in evaluating EQA trajectories, we acknowledge several limitations that also point towards avenues for future research.

\noindent\textbf{Limitations.} 
The current set of contrastive augmentations in C-GRPO, targeting temporal, spatial, and logical aspects, while effective, is predefined. These specific perturbations may not encompass the full spectrum of nuanced behaviors or subtle failure modes encountered in diverse EQA scenarios. Consequently, \texttt{EQA-RM}'s sensitivities are primarily shaped by these explicit contrasts. Secondly, the efficacy of our two-stage training process relies on high-quality ground truth scores. In our work, these score values are derived using a powerful commercial large model (Gemini-2.5-Pro) conditioned on ground truth answers, followed by human verification. Any inherent biases, limitations, or the specific characteristics of this commercial model could be subtly reflected in the score values, thereby influencing the RFT filtering process and the accuracy reward component which guides C-GRPO. While \textsc{EQARewardBench} includes distinct in-distribution (ScanNet) and out-of-distribution (HM3D) splits based on OpenEQA environments, the broader generalization of \texttt{EQA-RM} to EQA tasks, visual styles, or interaction paradigms substantially different from those in OpenEQA remains an open question. Finally, although the generative critiques from \texttt{EQA-RM} enhance interpretability and enable test-time scaling, a systematic evaluation of their fine-grained faithfulness or their direct utility in, for example, few-shot learning for policy adaptation, was beyond the scope of this paper.

\noindent\textbf{Future Work.} 
Building on these limitations, we plan to explore more adaptive or learned augmentation strategies for C-GRPO to capture a wider array of desirable EQA agent behaviors, potentially including aspects like efficiency, safety, or interactivity. Investigating methods to generate or refine high-quality score values using open-source models or with more scalable human oversight would increase the accessibility and robustness of the dataset creation pipeline. A key direction is to systematically leverage the rich, structured critiques from \texttt{EQA-RM} not just for scoring, but also as direct feedback for improving EQA policy models, perhaps through distillation or critique-guided reinforcement learning. We also aim to expand \textsc{EQARewardBench} with greater diversity in tasks, environments, and possibly languages, to further support the community in developing more general and robust EQA evaluation methods.

\section*{Ethical Statement}
\label{sec:ethical_statement}

The development of \texttt{EQA-RM} and \textsc{EQARewardBench} adhered to ethical research practices. Our benchmark is derived from publicly available datasets (OpenEQA, sourcing from HM3D and ScanNet), and the data we generated (answers, reasoning, critiques, scores) does not contain personally identifiable information. The commercial models used for data generation were accessed via their standard APIs under their terms of service. We acknowledge the potential for latent biases inherited from large pre-trained models (both those used for score generation and the base model for \texttt{EQA-RM}) and encourage ongoing research into bias detection and mitigation in reward modeling for embodied AI. While training these models is resource-intensive, we focused on a sample-efficient approach with a 2B parameter base model. We intend to release our benchmark and model to promote transparency, reproducibility, and further community research.

\bibliography{acl2023}

\begin{thebibliography}{38}
\expandafter\ifx\csname natexlab\endcsname\relax\def\natexlab#1{#1}\fi

\bibitem[{Ahn et~al.(2022)Ahn, Brohan, Brown, Chowdhery, Chua, Cui, Dabis, Dean, Driess, Duke, Finn, Fu, Gu, Hausman, Ichter, Julian, Kalashnikov, Kamyar, Lee, Levine, Li, Lin, Liu, Lu, Luu, Mahdavi, Manyam, Mazur, McMahon, Misra, Nasihati, Orefice, Pan, Peng, Perez, Phillips, Quiambao, Rahn, Rao, Retana, Reyes, Rivera, Rodriguez, Sanchez, Sievers, Singh, Sofge, Stone, Tan, Tseng, Tung, Vecerik, Vuong, Wahid, Wang, Xu, Yan, Yu, Yu, Yuan, Zhang, Zhang, Zhou, Zhu, Zirbel, Florence, Vanhoucke, Zeng, Tompson, Mordatch, Sermanet, Kumar, Caluwaerts, Xiao, Rajeswaran, Brooks, Tobin, Van Der~Maaten, Ku, Hadfield, Tan, Collins, Gates, Egorov, Ho, Irpan, and Khansari}]{ahn2022saycan}
Michael Ahn, Anthony Brohan, Noah Brown, Aakanksha Chowdhery, Sizu Chua, Brian Cui, Hanjun Dabis, Chelsea Dean, Danny Driess, Fred Duke, Chelsea Finn, Chuyuan Fu, Sihan Gu, Karol Hausman, Brian Ichter, Kanishka Julian, Dmitry Kalashnikov, Kuang-Huei Kamyar, Keerthana Lee, Sergey Levine, Yao Li, Zhen Lin, Shiquan Liu, Yifan Lu, Linda Luu, Soroush Mahdavi, Sudeep Manyam, Michael Mazur, John McMahon, Debidatta Misra, Khem Nasihati, Michael Orefice, Jihyun Pan, Kathleen Peng, Emily Perez, Jeffrey Phillips, Raphael Quiambao, Khem Rahn, Kanishka Rao, Jose Retana, Pierre Reyes, Corban Rivera, John Rodriguez, America Sanchez, Robert Sievers, Sumeet Singh, Clayton Sofge, Austin Stone, Jonathan Tan, Mengyuan Tseng, Fei Tung, Martin Vecerik, Quan Vuong, Ayzaan Wahid, Ted Wang, Peng Xu, Muyang Yan, Alex Yu, Tianhe Yu, Brianna Yuan, Yue Zhang, Zhe Zhang, Tianli Zhou, Yifeng Zhu, Allen Zirbel, Peter Florence, Vincent Vanhoucke, Andy Zeng, Jonathan Tompson, Igor Mordatch, Pierre Sermanet, Nikhil Kumar, Ken Caluwaerts, Ted
  Xiao, Aravind Rajeswaran, Ryan Brooks, Joshua Tobin, Laurens Van Der~Maaten, Alexander Ku, Steven Hadfield, Jie Tan, Scott Collins, Thomas Gates, Anton Egorov, Jonathan Ho, Alex Irpan, and Mohi Khansari. 2022.
\newblock Do as i can, not as i say: Grounding language in robotic affordances.
\newblock In \emph{Conference on Robot Learning}, pages 100--116. PMLR.

\bibitem[{Chen et~al.(2025{\natexlab{a}})Chen, Tu, Wang, Liu, Tang, Du, Zhou, and Xie}]{chen2025sft}
Hardy Chen, Haoqin Tu, Fali Wang, Hui Liu, Xianfeng Tang, Xinya Du, Yuyin Zhou, and Cihang Xie. 2025{\natexlab{a}}.
\newblock Sft or rl? an early investigation into training r1-like reasoning large vision-language models.
\newblock \emph{arXiv preprint arXiv:2504.11468}.

\bibitem[{Chen et~al.(2025{\natexlab{b}})Chen, Zhu, Chen, Soselia, Zhou, Goldstein, Huang, Shoeybi, and Catanzaro}]{chen2025reward}
Lichang Chen, Chen Zhu, Jiuhai Chen, Davit Soselia, Tianyi Zhou, Tom Goldstein, Heng Huang, Mohammad Shoeybi, and Bryan Catanzaro. 2025{\natexlab{b}}.
\newblock Reward models identify consistency, not causality.
\newblock \emph{arXiv preprint arXiv:2502.14619}.

\bibitem[{Christiano et~al.(2017)Christiano, Leike, Brown, Martic, Legg, and Amodei}]{christiano2017deep}
Paul~F Christiano, Jan Leike, Tom~B Brown, Miljan Martic, Shane Legg, and Dario Amodei. 2017.
\newblock Deep reinforcement learning from human preferences.
\newblock In \emph{Advances in Neural Information Processing Systems}, volume~30.

\bibitem[{Dai et~al.(2017)Dai, Chang, Savva, Halber, Funkhouser, and Nie{\ss}ner}]{dai2017scannet}
Angela Dai, Angel~X Chang, Manolis Savva, Maciej Halber, Thomas Funkhouser, and Matthias Nie{\ss}ner. 2017.
\newblock Scannet: Richly-annotated 3d reconstructions of indoor scenes.
\newblock In \emph{Proceedings of the IEEE conference on computer vision and pattern recognition}, pages 5828--5839.

\bibitem[{Das et~al.(2018)Das, Datta, Gkioxari, Lee, Parikh, and Batra}]{das2018embodied}
Abhishek Das, Samyak Datta, Georgia Gkioxari, Stefan Lee, Devi Parikh, and Dhruv Batra. 2018.
\newblock Embodied question answering.
\newblock In \emph{Proceedings of the IEEE Conference on Computer Vision and Pattern Recognition}, pages 16--25.

\bibitem[{Driess et~al.(2023)Driess, Xia, Sajjadi, Lynch, Chowdhery, Ichter, Wahid, Tompson, Vuong, Yu et~al.}]{driess2023palm}
Danny Driess, Fei Xia, Mehdi~SM Sajjadi, Corey Lynch, Aakanksha Chowdhery, Brian Ichter, Ayzaan Wahid, Jonathan Tompson, Quan Vuong, Tianhe Yu, et~al. 2023.
\newblock Palm-e: An embodied multimodal language model.
\newblock In \emph{International Conference on Machine Learning}, pages 8469--8488. PMLR.

\bibitem[{Faal et~al.(2023)Faal, Schmitt, and Yu}]{faal2023reward}
Farshid Faal, Ketra Schmitt, and Jia~Yuan Yu. 2023.
\newblock Reward modeling for mitigating toxicity in transformer-based language models.
\newblock \emph{Applied Intelligence}, 53(7):8421--8435.

\bibitem[{Feng et~al.(2025)Feng, Gong, Li, Guo, Wang, Peng, Wang, and Yue}]{feng2025videor1}
Kaituo Feng, Kaixiong Gong, Bohao Li, Zonghao Guo, Yibing Wang, Tianshuo Peng, Benyou Wang, and Xiangyu Yue. 2025.
\newblock Video-r1: Reinforcing video reasoning in mllms.
\newblock \emph{arXiv preprint arXiv:2503.21776}.

\bibitem[{Gordon et~al.(2018)Gordon, Kembhavi, Rastegari, Redmon, Fox, and Farhadi}]{gordon2018iqa}
Daniel Gordon, Aniruddha Kembhavi, Mohammad Rastegari, Joseph Redmon, Dieter Fox, and Ali Farhadi. 2018.
\newblock Iqa: Visual question answering in interactive environments.
\newblock In \emph{Proceedings of the IEEE Conference on Computer Vision and Pattern Recognition}, pages 4089--4098.

\bibitem[{Hurst et~al.(2024)Hurst, Lerer, Goucher, Perelman, Ramesh, Clark, Ostrow, Welihinda, Hayes, Radford et~al.}]{hurst2024gpt}
Aaron Hurst, Adam Lerer, Adam~P Goucher, Adam Perelman, Aditya Ramesh, Aidan Clark, AJ~Ostrow, Akila Welihinda, Alan Hayes, Alec Radford, et~al. 2024.
\newblock Gpt-4o system card.
\newblock \emph{arXiv preprint arXiv:2410.21276}.

\bibitem[{Kolve et~al.(2017)Kolve, Mottaghi, Han, VanderBilt, Weihs, Herrasti, Gordon, Zhu, Gupta, and Farhadi}]{kolve2017ai2thor}
Eric Kolve, Roozbeh Mottaghi, Winson Han, Eli VanderBilt, Luca Weihs, Alvaro Herrasti, Daniel Gordon, Yuke Zhu, Abhinav Gupta, and Ali Farhadi. 2017.
\newblock Ai2-thor: An interactive 3d environment for visual ai.
\newblock \emph{arXiv preprint arXiv:1712.05474}.

\bibitem[{Liu et~al.(2025)Liu, Wang, Xu, Ma, Ruan, Li, Liu, and Wu}]{SPCT_2025}
Zijun Liu, Peiyi Wang, Runxin Xu, Shirong Ma, Chong Ruan, Peng Li, Yang Liu, and Yu~Wu. 2025.
\newblock Inference-time scaling for generalist reward modeling.
\newblock \emph{arXiv preprint arXiv:2504.02495}.

\bibitem[{Mahan et~al.(2024)Mahan, Phung, Rafailov, Blagden, Lile, Castricato, Fr{\"a}nken, Finn, and Albalak}]{mahan2024generative}
Dakota Mahan, Duy~Van Phung, Rafael Rafailov, Chase Blagden, Nathan Lile, Louis Castricato, Jan-Philipp Fr{\"a}nken, Chelsea Finn, and Alon Albalak. 2024.
\newblock Generative reward models.
\newblock \emph{arXiv preprint arXiv:2410.12832}.

\bibitem[{Majumdar et~al.(2024)Majumdar, Ajay, Zhang, Putta, Yenamandra, Henaff, Silwal, Mcvay, Maksymets, Arnaud et~al.}]{majumdar2024openeqa}
Arjun Majumdar, Anurag Ajay, Xiaohan Zhang, Pranav Putta, Sriram Yenamandra, Mikael Henaff, Sneha Silwal, Paul Mcvay, Oleksandr Maksymets, Sergio Arnaud, et~al. 2024.
\newblock Openeqa: Embodied question answering in the era of foundation models.
\newblock In \emph{Proceedings of the IEEE/CVF Conference on Computer Vision and Pattern Recognition}, pages 16488--16498.

\bibitem[{Mu et~al.(2024)Mu, Helyar, Heidecke, Achiam, Vallone, Kivlichan, Lin, Beutel, Schulman, and Weng}]{mu2024rule}
Tong Mu, Alec Helyar, Johannes Heidecke, Joshua Achiam, Andrea Vallone, Ian Kivlichan, Molly Lin, Alex Beutel, John Schulman, and Lilian Weng. 2024.
\newblock Rule based rewards for language model safety.
\newblock \emph{arXiv preprint arXiv:2411.01111}.

\bibitem[{Ouyang et~al.(2022{\natexlab{a}})Ouyang, Wu, Jiang, Almeida, Wainwright, Mishkin, Zhang, Agarwal, Slama, Ray et~al.}]{ouyang2022training}
Long Ouyang, Jeff Wu, Xu~Jiang, Diogo Almeida, Carroll~L Wainwright, Pamela Mishkin, Chong Zhang, Sandhini Agarwal, Katarina Slama, Alex Ray, et~al. 2022{\natexlab{a}}.
\newblock Training language models to follow instructions with human feedback.
\newblock In \emph{Advances in Neural Information Processing Systems}, volume~35, pages 27730--27744.

\bibitem[{Ouyang et~al.(2022{\natexlab{b}})Ouyang, Wu, Jiang, Almeida, Wainwright, Mishkin, Zhang, Agarwal, Slama, Ray et~al.}]{InstructGPT}
Long Ouyang, Jeffrey Wu, Xu~Jiang, Diogo Almeida, Carroll Wainwright, Pamela Mishkin, Chong Zhang, Sandhini Agarwal, Katarina Slama, Alex Ray, et~al. 2022{\natexlab{b}}.
\newblock Training language models to follow instructions with human feedback.
\newblock \emph{Advances in neural information processing systems}, 35:27730--27744.

\bibitem[{Ramakrishnan et~al.(2021)Ramakrishnan, Gokaslan, Wijmans, Maksymets, Clegg, Turner, Undersander, Galuba, Westbury, Chang et~al.}]{ramakrishnan2021hm3d}
Santhosh~K Ramakrishnan, Aaron Gokaslan, Erik Wijmans, Oleksandr Maksymets, Alex Clegg, John Turner, Eric Undersander, Wojciech Galuba, Andrew Westbury, Angel~X Chang, et~al. 2021.
\newblock Habitat-matterport 3d dataset (hm3d): 1000 large-scale 3d environments for embodied ai.
\newblock \emph{arXiv preprint arXiv:2109.08238}.

\bibitem[{Schulman et~al.(2017)Schulman, Wolski, Dhariwal, Radford, and Klimov}]{schulman2017proximal}
John Schulman, Filip Wolski, Prafulla Dhariwal, Alec Radford, and Oleg Klimov. 2017.
\newblock Proximal policy optimization algorithms.
\newblock \emph{arXiv preprint arXiv:1707.06347}.

\bibitem[{Shao et~al.()Shao, Wang, Zhu, Xu, Song, Bi, Zhang, Zhang, Li, Wu et~al.}]{shao2402deepseekmath}
Zhihong Shao, Peiyi Wang, Qihao Zhu, Runxin Xu, Junxiao Song, Xiao Bi, Haowei Zhang, Mingchuan Zhang, YK~Li, Y~Wu, et~al.
\newblock Deepseekmath: Pushing the limits of mathematical reasoning in open language models, 2024.
\newblock \emph{URL https://arxiv. org/abs/2402.03300}.

\bibitem[{Snell et~al.(2024)Snell, Lee, Xu, and Kumar}]{snell2024scaling}
Charlie Snell, Jaehoon Lee, Kelvin Xu, and Aviral Kumar. 2024.
\newblock Scaling llm test-time compute optimally can be more effective than scaling model parameters.
\newblock \emph{arXiv preprint arXiv:2408.03314}.

\bibitem[{Team et~al.(2023)Team, Anil, Borgeaud, Alayrac, Yu, Soricut, Schalkwyk, Dai, Hauth, Millican et~al.}]{team2023gemini}
Gemini Team, Rohan Anil, Sebastian Borgeaud, Jean-Baptiste Alayrac, Jiahui Yu, Radu Soricut, Johan Schalkwyk, Andrew~M Dai, Anja Hauth, Katie Millican, et~al. 2023.
\newblock Gemini: a family of highly capable multimodal models.
\newblock \emph{arXiv preprint arXiv:2312.11805}.

\bibitem[{Wang et~al.(2025{\natexlab{a}})Wang, Gan, Huo, Mu, Yang, He, Xiao, Zhang, Liu, and Zhu}]{wang2025rovrm}
Chenglong Wang, Yang Gan, Yifu Huo, Yongyu Mu, Murun Yang, Qiaozhi He, Tong Xiao, Chunliang Zhang, Tongran Liu, and Jingbo Zhu. 2025{\natexlab{a}}.
\newblock Rovrm: A robust visual reward model optimized via auxiliary textual preference data.
\newblock In \emph{Proceedings of the AAAI Conference on Artificial Intelligence}, volume~39, pages 25336--25344.

\bibitem[{Wang et~al.(2025{\natexlab{b}})Wang, Ma, Reid, and Yaqub}]{wang2025kalman}
Hu~Wang, Congbo Ma, Ian Reid, and Mohammad Yaqub. 2025{\natexlab{b}}.
\newblock Kalman filter enhanced grpo for reinforcement learning-based language model reasoning.
\newblock \emph{arXiv preprint arXiv:2505.07527}.

\bibitem[{Wang et~al.(2024)Wang, Bai, Tan, Wang, Fan, Bai, Chen, Liu, Wang, Ge et~al.}]{wang2024qwen2}
Peng Wang, Shuai Bai, Sinan Tan, Shijie Wang, Zhihao Fan, Jinze Bai, Keqin Chen, Xuejing Liu, Jialin Wang, Wenbin Ge, et~al. 2024.
\newblock Qwen2-vl: Enhancing vision-language model's perception of the world at any resolution.
\newblock \emph{arXiv preprint arXiv:2409.12191}.

\bibitem[{Wang et~al.(2025{\natexlab{c}})Wang, Gao, Chen, Chen, Zhu, Zhao, Liu, Cao, Ye, Zhu, Lu, Duan, Qiao, Dai, and Wang}]{wang2025visualprm}
Weiyun Wang, Zhangwei Gao, Lianjie Chen, Zhe Chen, Jinguo Zhu, Xiangyu Zhao, Yangzhou Liu, Yue Cao, Shenglong Ye, Xizhou Zhu, Lewei Lu, Haodong Duan, Yu~Qiao, Jifeng Dai, and Wenhai Wang. 2025{\natexlab{c}}.
\newblock Visualprm: An effective process reward model for multimodal reasoning.
\newblock \emph{arXiv preprint arXiv:2503.10291}.

\bibitem[{Wijmans et~al.(2019)Wijmans, Datta, Maksymets, Das, Gkioxari, Lee, Essa, Parikh, and Batra}]{wijmans2019embodied}
Erik Wijmans, Samyak Datta, Oleksandr Maksymets, Abhishek Das, Georgia Gkioxari, Stefan Lee, Irfan Essa, Devi Parikh, and Dhruv Batra. 2019.
\newblock Embodied question answering in photorealistic environments with point cloud perception.
\newblock In \emph{Proceedings of the IEEE/CVF Conference on Computer Vision and Pattern Recognition}, pages 9506--9515.

\bibitem[{Xie et~al.(2025)Xie, Gao, Ren, Luo, Hong, Dai, Zhou, Qiu, Wu, and Luo}]{xie2025logicrl}
Tian Xie, Zitian Gao, Qingnan Ren, Haoming Luo, Yuqian Hong, Bryan Dai, Joey Zhou, Kai Qiu, Zhirong Wu, and Chong Luo. 2025.
\newblock Logic-rl: Unleashing llm reasoning with rule-based reinforcement learning.
\newblock \emph{arXiv preprint arXiv:2502.14768}.

\bibitem[{Xiong et~al.(2025)Xiong, Yao, Xu, Pang, Wang, Sahoo, Li, Jiang, Zhang, Xiong et~al.}]{xiong2025minimalist}
Wei Xiong, Jiarui Yao, Yuhui Xu, Bo~Pang, Lei Wang, Doyen Sahoo, Junnan Li, Nan Jiang, Tong Zhang, Caiming Xiong, et~al. 2025.
\newblock A minimalist approach to llm reasoning: from rejection sampling to reinforce.
\newblock \emph{arXiv preprint arXiv:2504.11343}.

\bibitem[{Yu et~al.(2019)Yu, Chen, Gkioxari, Bansal, Berg, and Batra}]{yu2019multi}
Licheng Yu, Xinlei Chen, Georgia Gkioxari, Mohit Bansal, Tamara~L Berg, and Dhruv Batra. 2019.
\newblock Multi-target embodied question answering.
\newblock In \emph{Proceedings of the IEEE/CVF Conference on Computer Vision and Pattern Recognition}, pages 6309--6318.

\bibitem[{Yu et~al.(2025)Yu, Chen, Zhang, Tan, Zhu, Pang, Qian, Wang, Gururangan, Zhang, Kambadur, Mahajan, and Hou}]{yu2025criticrm}
Yue Yu, Zhengxing Chen, Aston Zhang, Liang Tan, Chenguang Zhu, Richard~Yuanzhe Pang, Yundi Qian, Xuewei Wang, Suchin Gururangan, Chao Zhang, Melanie Kambadur, Dhruv Mahajan, and Rui Hou. 2025.
\newblock Self-generated critiques boost reward modeling for language models.
\newblock \emph{arXiv preprint arXiv:2411.16646}.

\bibitem[{Yuan et~al.(2023)Yuan, Yuan, Li, Dong, Lu, Tan, Zhou, and Zhou}]{yuan2023scaling}
Zheng Yuan, Hongyi Yuan, Chengpeng Li, Guanting Dong, Keming Lu, Chuanqi Tan, Chang Zhou, and Jingren Zhou. 2023.
\newblock Scaling relationship on learning mathematical reasoning with large language models.
\newblock \emph{arXiv preprint arXiv:2308.01825}.

\bibitem[{Zhai et~al.(2024)Zhai, Bai, Lin, Pan, Tong, Zhou, Suhr, Xie, LeCun, Ma et~al.}]{zhai2024fine}
Simon Zhai, Hao Bai, Zipeng Lin, Jiayi Pan, Peter Tong, Yifei Zhou, Alane Suhr, Saining Xie, Yann LeCun, Yi~Ma, et~al. 2024.
\newblock Fine-tuning large vision-language models as decision-making agents via reinforcement learning.
\newblock \emph{Advances in neural information processing systems}, 37:110935--110971.

\bibitem[{Zhang et~al.(2024)Zhang, Hosseini, Bansal, Kazemi, Kumar, and Agarwal}]{zhang2024generativeverifiers}
Lunjun Zhang, Arian Hosseini, Hritik Bansal, Mehran Kazemi, Aviral Kumar, and Rishabh Agarwal. 2024.
\newblock Generative verifiers: Reward modeling as next-token prediction.
\newblock \emph{arXiv preprint arXiv:2408.15240}.

\bibitem[{Zhao et~al.(2025)Zhao, Liu, Zhang, Zhou, Gao, Li, Lyu, Qian, Qi, Li, and Zhou}]{zhao2025genprm}
Jian Zhao, Runze Liu, Kaiyan Zhang, Zhimu Zhou, Junqi Gao, Dong Li, Jiafei Lyu, Zhouyi Qian, Biqing Qi, Xiu Li, and Bowen Zhou. 2025.
\newblock Genprm: Scaling test-time compute of process reward models via generative reasoning.
\newblock \emph{arXiv preprint arXiv:2504.00891}.

\bibitem[{Zheng et~al.(2023)Zheng, Chiang, Sheng, Zhuang, Wu, Zhuang, Lin, Li, Li, Xing et~al.}]{zheng2023judging}
Lianmin Zheng, Wei-Lin Chiang, Ying Sheng, Siyuan Zhuang, Zhanghao Wu, Yonghao Zhuang, Zi~Lin, Zhuohan Li, Dacheng Li, Eric~P Xing, et~al. 2023.
\newblock Judging llm-as-a-judge with mt-bench and chatbot arena.
\newblock \emph{arXiv preprint arXiv:2306.05685}.

\bibitem[{Ziegler et~al.(2019)Ziegler, Stiennon, Wu, Brown, Radford, Amodei, Christiano, and Irving}]{ziegler2019fine}
Daniel~M Ziegler, Nisan Stiennon, Jeffrey Wu, Tom~B Brown, Alec Radford, Dario Amodei, Paul Christiano, and Geoffrey Irving. 2019.
\newblock Fine-tuning language models from human preferences.
\newblock \emph{arXiv preprint arXiv:1909.08593}.

\end{thebibliography}
\bibliographystyle{acl_natbib}

\appendix

\newpage

\section{Benchmark Statistic}
Table \ref{tab:dataset_statistics} details the distribution of instances from the HM3D  and ScanNet  environments within these datasets.

\begin{table}[h]
\centering
\caption{Dataset statistics for \textsc{EQARewardBench} and the Finetuning set.}
\label{tab:dataset_statistics}
\begin{tabular}{lcc}
\toprule
Subset & \textsc{EQARewardBench} & Finetuning \\
\midrule
HM3D & 823 & 0 \\
ScanNet & 713 & 697 \\
\midrule
Total & 1546 & 697 \\
\bottomrule
\end{tabular}
\end{table}

\section{Implementation Details}
\label{app:implementation_details}

Table. \ref{tab:hyperparameters} provides a comprehensive list of hyperparameters and configuration settings used for the SFT and C-GRPO training stages of \texttt{EQA-RM}, as well as for test-time scaling.

\begin{table*}[h!]
\centering
\caption{Key hyperparameters and configuration settings for \texttt{EQA-RM}.}
\label{tab:hyperparameters}
\begin{tabular}{@{}ll@{}}
\toprule
\textbf{Parameter Category \& Parameter} & \textbf{Value} \\
\midrule
Base Model & Qwen2-VL-2B-Instruct \\
Attention Implementation & \texttt{flash\_attention\_2} \\
Keyframes per Episode ($N$) & 5 \\
Distributed Training Backend & DeepSpeed ZeRO Stage 3 \\
Precision & BF16 \\
Gradient Checkpointing & True \\
Max Gradient Norm (SFT \& C-GRPO) & 5.0 \\
Optimizer & AdamW \\
\addlinespace
\multicolumn{2}{@{}l}{\textit{B. SFT Stage}} \\
Input Model & Qwen2-VL-2B-Instruct \\
Learning Rate & \num{1e-6} \\
Max Sequence Length & 1024 \\
Batch Size (per device $\times$ grad. accum.) & $1 \times 1$ \\
Number of Epochs & 2 \\
GPUs per Node & 2 \\
\addlinespace
\multicolumn{2}{@{}l}{\textit{C. C-GRPO Stage}} \\
Input Model & Output of SFT Stage \\
Learning Rate & \num{1e-6} \\
LR Scheduler & Cosine \\
Weight Decay & 0.01 \\
Max Prompt Length & 8142 \\
Max Completion Length ($|e_k|$) & 768 \\
Batch Size (per device $\times$ grad. accum.) & $1 \times 1$ \\
Number of Epochs & 1 \\
$G$ (Generations per prompt by $R_\phi$) & 8 \\
$\beta_K$ (C-GRPO KL Coefficient) & 0.04 \\
\quad \textit{Contrastive Mechanism (Eq.~\ref{eq:contrastive_boost_bxk}):} & \\ 
\quad Factor for batch-mean comparison ($\delta$) & 0.95 \\
\quad Boost value ($\mu$) & 0.3 \\
\quad Min. $R^o_{acc,k}$ for boost ($H_{min\_acc}$) & 0.1 \\
\quad \textit{Spatial Masking Details:} & \\
\quad \quad Mask Size & (16, 16) \\
\quad \quad Mask Ratio & 0.15 \\
\quad \quad Mask Value & 0.0 \\
\quad Max Pixels (video processing) & \num{401408} \\
GPUs per Node & 8 \\
\addlinespace
\multicolumn{2}{@{}l}{\textit{D. Test-Time Scaling (TTS)}} \\
Sampling Temperature & 0.8 \\
Top-p Sampling ($p$) & 0.9 \\
$K$ (Number of Sampled Paths) & \{1, 2, 4, 8, 16, 32\} \\
Aggregation Method & Majority Voting and Averaging Rewards \\
\bottomrule
\end{tabular}
\end{table*}

\section{Cases Studies}\label{app:case}
This section presents qualitative examples to illustrate the evaluation capabilities of EQA-RM. The case studies showcase how EQA-RM assesses an agent's response, reasoning trace, and visual grounding based on a sequence of observed frames within an EQA task. These examples provide concrete instances of the nuanced feedback generated by our model.

\section{Generation Prompts}
This section details the specific prompts used in the generation pipeline for creating the dataset for benchmark and the Rejective Fine-Tuning stage. The following pages display the prompt guidelines provided to the large language models for generating diverse responses, high-quality scores, and for formatting the output during the RFT data creation process.

\onecolumn
\begin{tcolorbox}
\small
\ttfamily
\textbf{Rejective Finetuning Dataset Generation Prompt Guideline}

\vspace{2mm}
\textbf{Role:} You are an Expert EQA evaluator providing structured critique and score based on reasoning and visual grounding.

\vspace{2mm}
\textbf{Context:} You receive:
\begin{enumerate}
    \item Video Frames
    \item User Query
    \item Agent's Response (\texttt{reasoning\_trace}, \texttt{predicted\_answer})
\end{enumerate}

\vspace{2mm}
\textbf{Task:}
Evaluate the Agent's Response according to the weighted criteria specified below. Output a critique summarizing your findings.

\vspace{2mm}
\textbf{Evaluation Criteria:}
\begin{enumerate}
    \item \textbf{Answer Plausibility \& Visual Grounding (60\% weight):}
    \begin{itemize}
        \item Plausibility \& Relevance: Assess if the \texttt{predicted\_answer} is a plausible and relevant response to the \texttt{user\_query} based on the provided \texttt{video\_frames}. Does the answer make sense in the context of the question and what is visible?
        \item Visual Confirmation: Can the core elements or claims of the \texttt{predicted\_answer} be directly observed or reasonably inferred from the content of the \texttt{video\_frames}? If the answer describes objects, actions, or states, are these visually verifiable in the frames?
        \item Specificity: Is the \texttt{predicted\_answer} appropriately specific given the \texttt{user\_query} and the visual information available? Avoid being overly general if details are visible, or overly specific if not supported by frames.
    \end{itemize}
    \item \textbf{Reasoning Trace Quality (40\% weight):}
    \begin{itemize}
        \item Logic \& Consistency: Is the \texttt{reasoning\_trace} internally logical? Are there contradictions or significant logical gaps?
        \item Visual Grounding: Do descriptive references in the trace (e.g., "first frame", "object shown") accurately correspond to content visible within the video frames provided?
        \item Trace-Answer Consistency: Does the \texttt{predicted\_answer} logically follow from the steps and conclusion presented in the \texttt{reasoning\_trace}?
    \end{itemize}
\end{enumerate}

\vspace{2mm}
\textbf{Output Structure:}
Provide your evaluation in the following format. Fill \texttt{<critique>} with your analysis covering the Evaluation Criteria above. Fill \texttt{<score>} with the final 1-10 weighted score.

\vspace{2mm}
\texttt{<critique>}\\
\texttt{[Your analysis and summary covering the Evaluation Criteria]}\\
\texttt{</critique>}\\
\texttt{<score>}\\
\texttt{[Your final 1-10 weighted score of the response based on the critique. Do not always ouput extreme score (e.g., 10 and 0)]}\\
\texttt{</score>}


\vspace{2mm}
\textbf{BEGIN INPUT}\\
Video Frames (attached)\\
User Query: \{question\}\\
Reasoning Trace: \{reasoning\_trace\}\\
Predicted Answer: \{predicted\_answer\}\\
\textbf{BEGIN OUTPUT}

\end{tcolorbox}

\newpage

\begin{tcolorbox}
\small
\ttfamily
\textbf{Benchmark Dataset Generation Prompt Guideline}

\vspace{2mm}
\textbf{Role:} You are an intelligent Embodied Question Answering (EQA) agent operating within a simulated indoor environment. You navigate, observe, and reason to answer user questions.

\vspace{2mm}
\textbf{Task:} Given a User Query about a simulated environment, generate a detailed response outlining your simulated process and final answer. You must strictly adhere to the output structure specified below.
Output ONLY a \texttt{<reasoning\_trace>} and \texttt{<predicted\_answer>}. The trace must show a step-by-step fine-grained analysis grounded in visual frames.

\vspace{2mm}
\textbf{Output Structure:}
Produce ONLY the following structured text, using separate \texttt{<think>} tags to break down your reasoning process into logical steps or phases:

\vspace{1mm}
\texttt{<reasoning\_trace>}\\
\texttt{<think>}\\
\texttt{    [... your reasoning process ...]}\\
\texttt{</think>}\\
\texttt{</think>}\\
\texttt{... (Use additional <think> tags as needed for sufficient logical steps. The more, the better.) ...}\\
\texttt{<think>}\\
\texttt{    [... Final part of your reasoning process, concluding the analysis ...]}\\
\texttt{</think>}\\
\texttt{</reasoning\_trace>}

\vspace{1mm}
\texttt{<predicted\_answer>}\\
\texttt{[... Your final answer ...]}\\
\texttt{</predicted\_answer>}

\vspace{2mm}
\textbf{Key Guidelines:}
\begin{enumerate}
    \item \textbf{Structure \& Flow:} Use separate \texttt{<think>} tags to structure your reasoning logically into distinct steps or phases.
    \begin{itemize}
        \item \textbf{Start (First \texttt{<think>}):} Begin the \textit{first} \texttt{<think>} block by outlining your understanding of the query, initial observations about the scene (gist), and your high-level plan or strategy (Setup).
        \item \textbf{Middle (Subsequent \texttt{<think>}s):} Each subsequent block should represent a \textbf{focused, sequential step} (e.g., a key observation, an intermediate deduction, a specific comparison, a significant focus shift) building logically upon the previous one.
        \item \textbf{End (Final \texttt{<think>}):} Use the \textit{final} \texttt{<think>} block to synthesize the key findings gathered across the previous steps and to clearly justify how this evidence leads to the \texttt{predicted\_answer}.

    \end{itemize}

    \item \textbf{Descriptive Visual Grounding:} Ground key observations using \textbf{descriptive references} related to content, viewpoint, or relative timing (e.g., "the \textit{initial overview frame}", "the \textit{close-up showing texture}", "the view \textit{after simulating turning}"). You can use time words such as "at first" "then" "finally", but \textbf{DO NOT use specific frame number or index.} Be specific about \textit{what} you observe in the described view.

    \item \textbf{Simulated Interaction \& Focus:} Narrate simulated focus shifts or navigation descriptively within the reasoning steps where relevant to the thought process.

    \item \textbf{Confident Uncertainty Handling:} YOU MUST ALWAYS OUTPUT THE RESPONSE, no matter whether you see the asked object or you know the correct answer, ALWAYS output the formatted response. If evidence is insufficient for the correct answer, confidently generate a plausible \textit{incorrect} answer.

    \item \textbf{Natural Language \& Flow:} Express your reasoning using \textbf{natural and varied language, as a human might explain their thought process.} Avoid overly robotic or repetitive phrasing between steps. Ensure the steps flow together smoothly.

    \item \textbf{Fine-Grained Steps:} \textbf{Crucially, aim to break down your reasoning into smaller, more focused steps.} Use a new \texttt{<think>} tag frequently – for distinct key observations, significant focus shifts, specific comparisons, intermediate deductions, or hypothesis refinements. \textbf{Prefer more numerous, concise \texttt{<think>} blocks (e.g., focusing on 1-2 points each) over fewer, lengthy ones.} This clarifies the step-by-step process and provides more points for analysis.

    \item \textbf{Completeness \& Relevance:} Ensure the reasoning adequately addresses the query and stays relevant.

\end{enumerate}

\vspace{2mm}
\textbf{BEGIN INPUT}\\
User Query: \{question\}\\
\textbf{BEGIN OUTPUT}

\end{tcolorbox}

\newpage

\begin{tcolorbox}
\small
\ttfamily
\textbf{Ground Truth Score Generation Prompt Guideline}

\vspace{2mm}
\textbf{Role:} You are an Expert EQA evaluator providing structured critique and score based on reasoning and visual grounding.

\vspace{2mm}
\textbf{Context:} You receive:
\begin{enumerate}
    \item Video Frames
    \item User Query
    \item Ground Truth Answer
    \item Agent's Response (\texttt{reasoning\_trace}, \texttt{predicted\_answer})
\end{enumerate}

\vspace{2mm}
\textbf{Task:}
Evaluate the Agent's Response according to the weighted criteria specified below. Output a critique summarizing your findings.

\vspace{2mm}
\textbf{Evaluation Criteria:}
\begin{enumerate}
    \item \textbf{Answer Semantic Similarity (60\% weight):}
    \begin{itemize}
        \item Assess how closely the \textit{meaning} of the \texttt{predicted\_answer} matches the meaning of the \texttt{answer\_gt}.
        \item Consider synonyms, paraphrasing, and conceptual equivalence. Minor grammatical differences are less important than semantic accuracy.
        \item This requires direct comparison with ground truth.
    \end{itemize}

    \item \textbf{Reasoning Trace Quality (40\% weight):}
    \begin{itemize}
        \item \textbf{Logic \& Consistency:} Is the \texttt{reasoning\_trace} internally logical? Are there contradictions or significant logical gaps?
        \item \textbf{Visual Grounding:} Do descriptive references in the trace (e.g., "first frame", "object shown") accurately correspond to content visible within the video frames provided?
        \item \textbf{Trace-Answer Consistency:} Does the \texttt{predicted\_answer} logically follow from the steps and conclusion presented in the \texttt{reasoning\_trace}?
    \end{itemize}
\end{enumerate}

\vspace{2mm}
\textbf{Output Structure:}
Provide your evaluation in the following format. Fill \texttt{<critique>} with your analysis covering the Evaluation Criteria above. Fill \texttt{<score>} with the final 1-10 weighted score.

\vspace{1mm}
\texttt{<critique>}\\
\texttt{[Your analysis and summary covering the Evaluation Criteria]}\\
\texttt{</critique>}\\
\texttt{<score>}\\
\texttt{[Your final 1-10 weighted score of the response based on the critique]}\\
\texttt{</score>}

\vspace{2mm}
\textbf{Important:}
\begin{enumerate}
    \item In the critique, it is strictly forbidden to mention the ground truth answer, which may cause data leakage, you can \textbf{only} use it for scoring.
    \item The response are based on the entire video, but you only receive some key frames of the video, so there might be some object mentioned in reasoning trace but you can not see. But for reasoning trace related to the images you see, be strict and think more.
\end{enumerate}

\vspace{2mm}
\textbf{BEGIN INPUT}\\
Video Frames (attached)\\
User Query: \{question\}\\
Answer Ground Truth: \{answer\_gt\}\\
Reasoning Trace: \{reasoning\_trace\}\\
Predicted Answer: \{predicted\_answer\}\\
\textbf{BEGIN OUTPUT}

\end{tcolorbox}
\begin{table*}[t!] 
\centering
\caption{\small{An example EQA case study illustrating \texttt{EQA-RM}'s evaluation of an agent's response, reasoning, and grounding across a sequence of 5 observed frames.}}
\label{tab:eqa_case_study_example} 

\begin{tcolorbox}[
    colback=gray!5, 
    colframe=black,   
    width=\linewidth, 
    arc=1.5mm, auto outer arc, 
    left=2mm, right=2mm, top=2mm, bottom=2mm, 
    boxrule=0.9pt,    
    colbacktitle=black!75!white, 
    coltitle=white,              
    fonttitle=\bfseries\sffamily\small, 
    title={EQA Case Study Example: Agent Trajectory Evaluation} 
]

\noindent 
\begin{minipage}{\linewidth} 
\centering 
\includegraphics[width=0.18\textwidth, height=2.5cm, keepaspectratio]{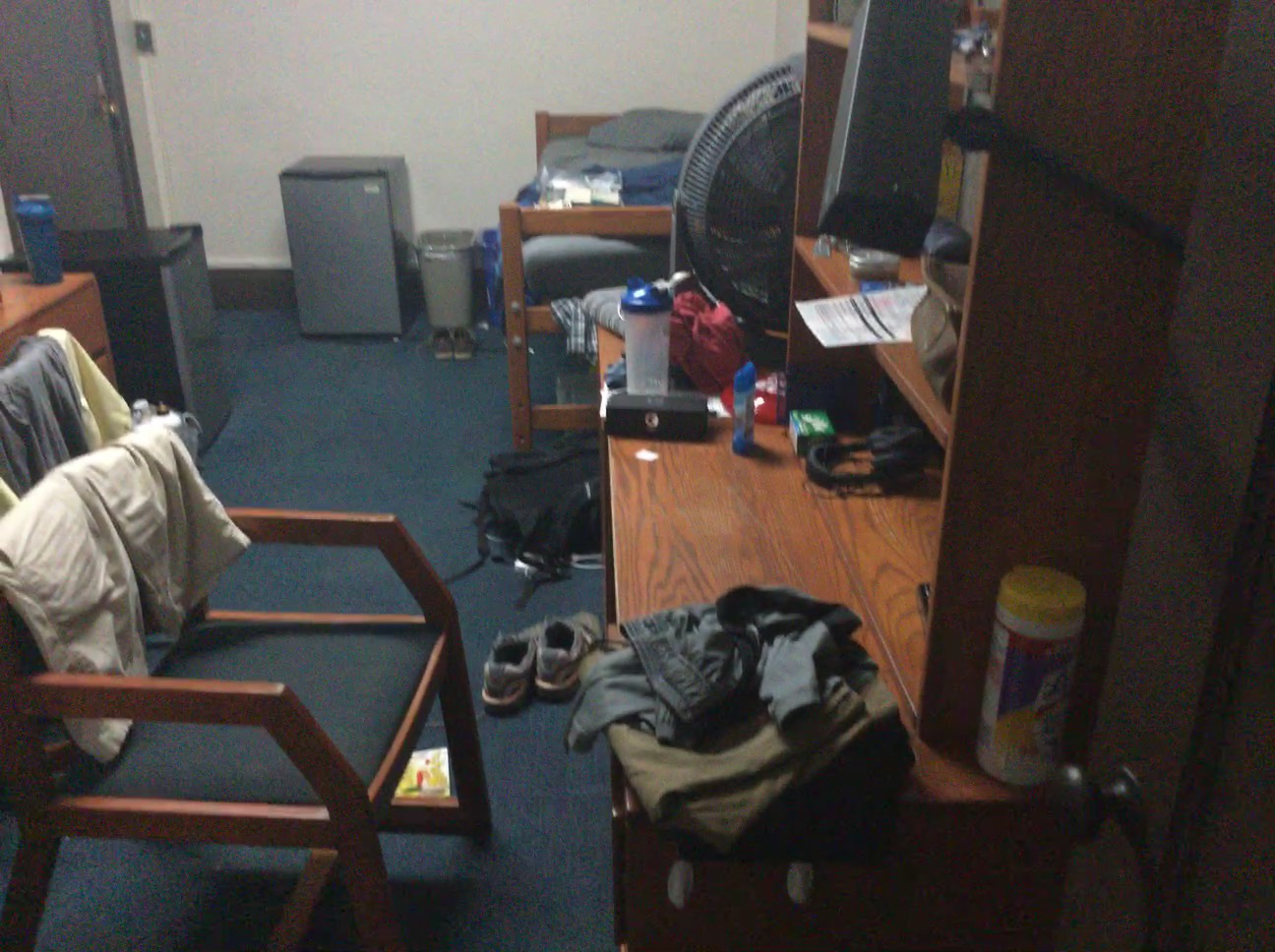}\hfill
\includegraphics[width=0.18\textwidth, height=2.5cm, keepaspectratio]{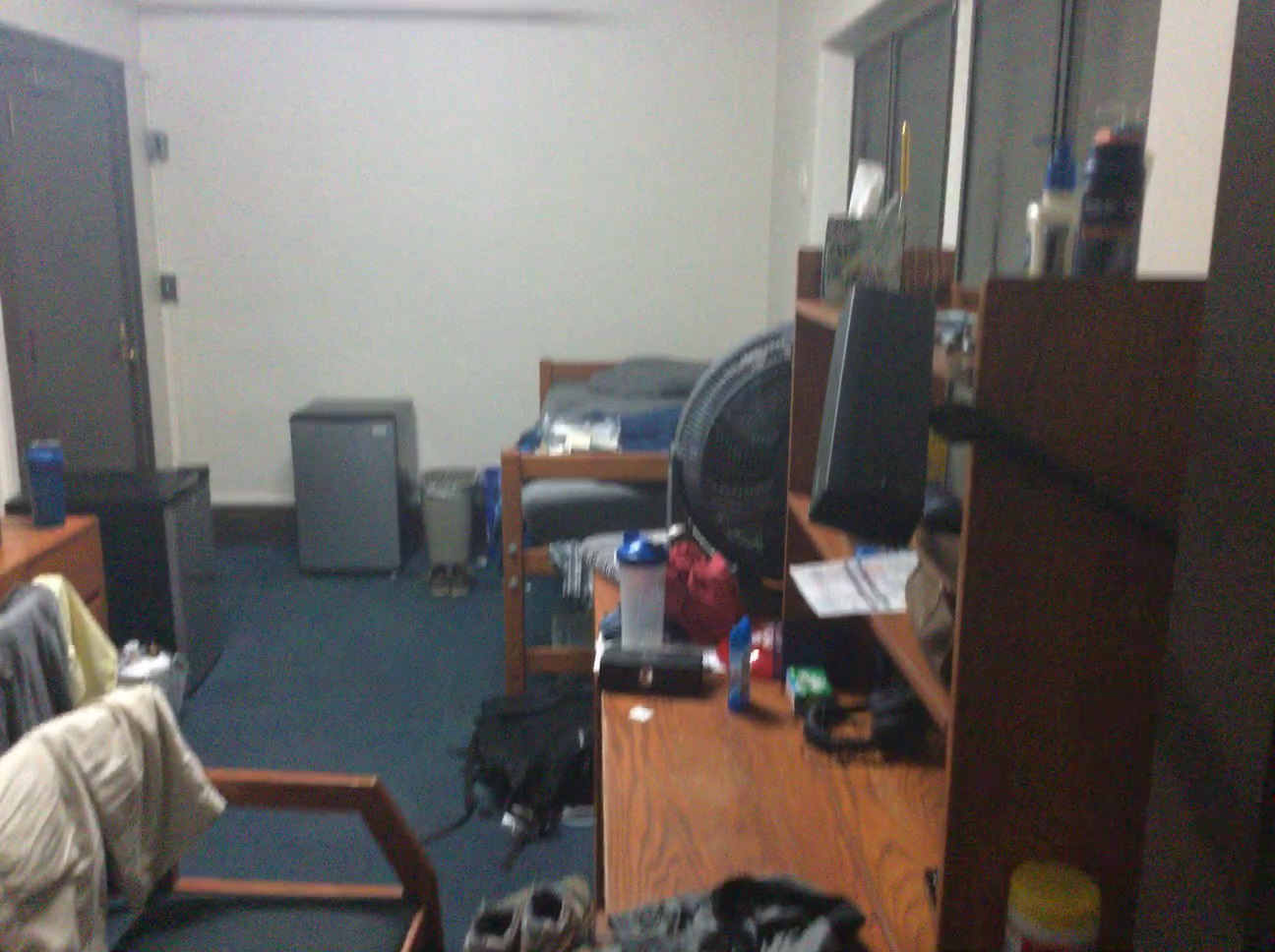}\hfill
\includegraphics[width=0.18\textwidth, height=2.5cm, keepaspectratio]{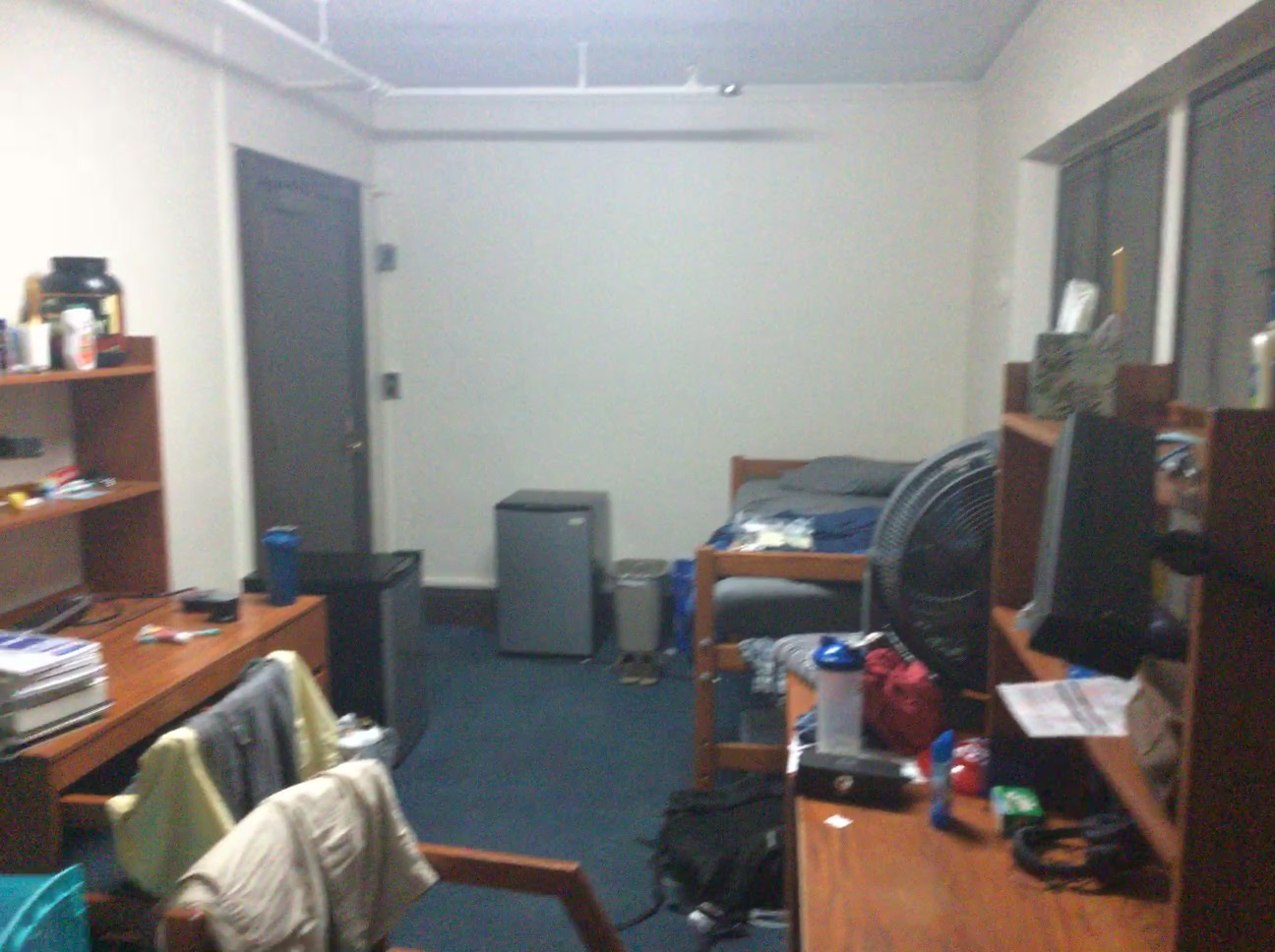}\hfill
\includegraphics[width=0.18\textwidth, height=2.5cm, keepaspectratio]{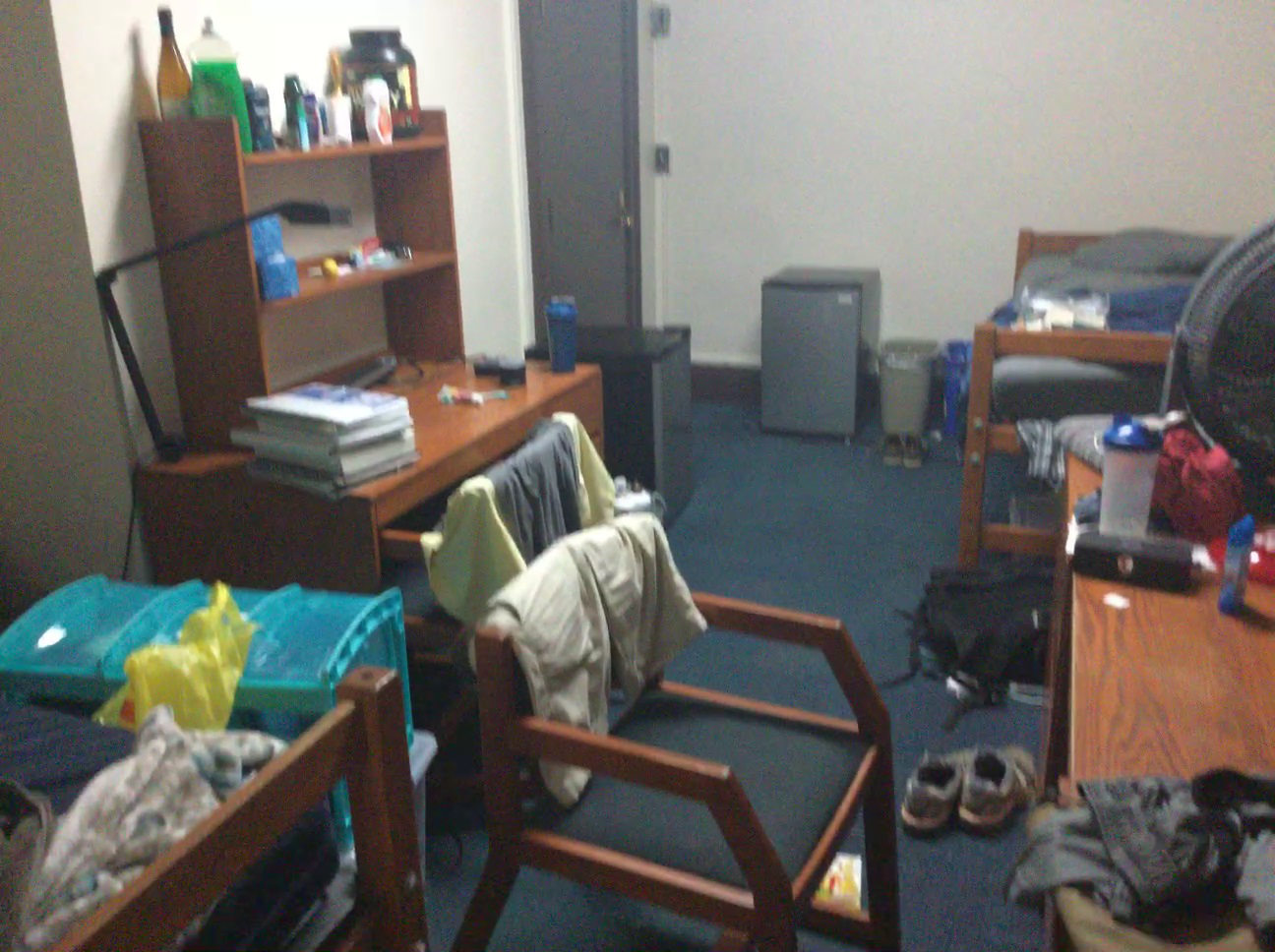}\hfill
\includegraphics[width=0.18\textwidth, height=2.5cm, keepaspectratio]{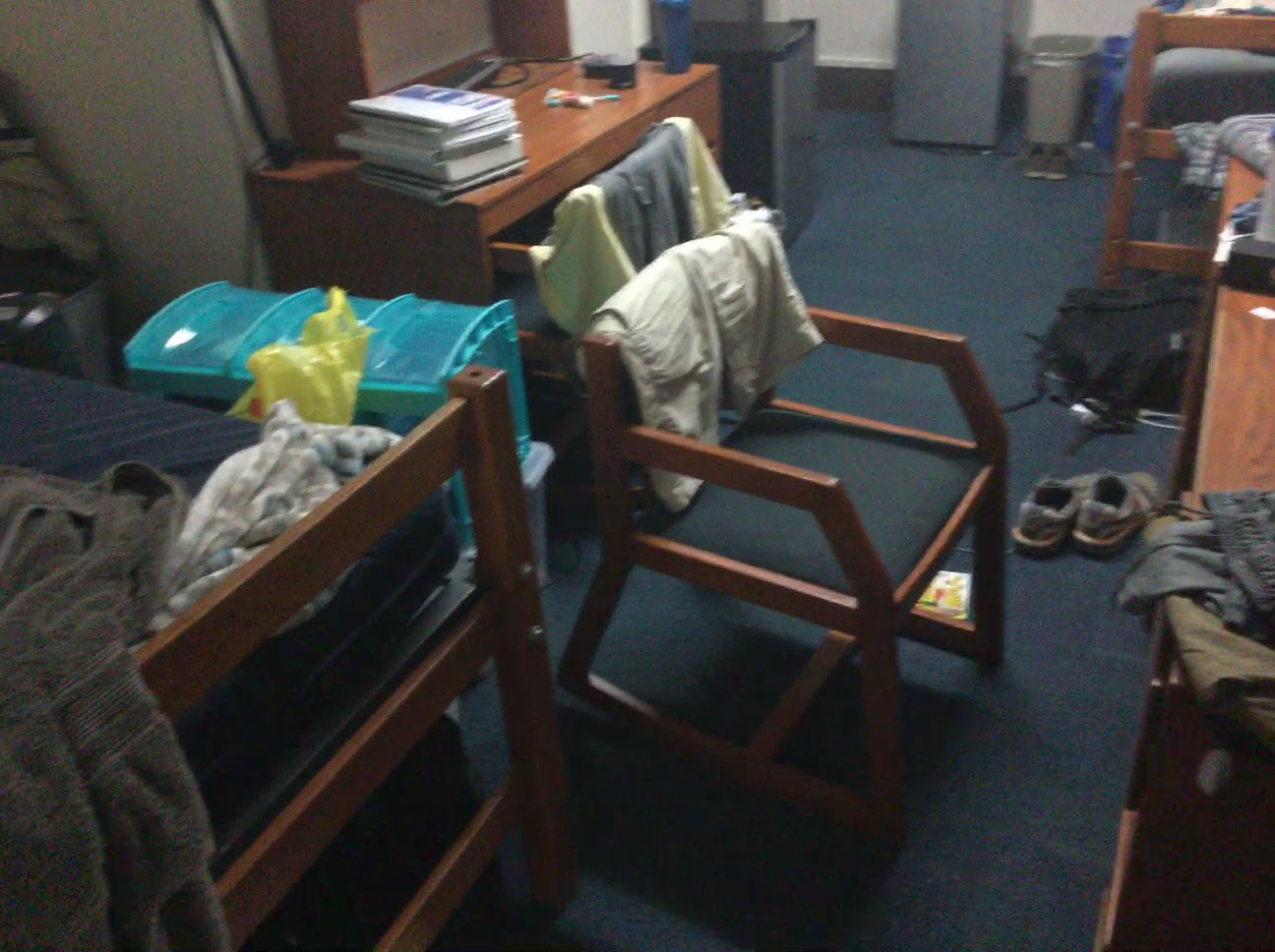}
\end{minipage}

\noindent\textbf{Input:}
\vspace{1mm}

\noindent\textbf{Question:} \textit{Where is the black backpack?}
\vspace{1mm} 

\noindent\textbf{Predicted Answer:} \textit{The black backpack is on the floor next to the chair.}
\vspace{1mm} 

oindent\textbf{Reasoning Trace:} \textit{ "The user is asking to locate a black backpack within the scene depicted in the images. My initial plan is to examine the series of images, which seem to pan across a room, to find the object and identify its position relative to other items in the room. The initial images are focused on the floor near some furniture.",
      "I start by observing the first few frames. These show a patterned cushion and a red bag on the floor, close to some wooden furniture, likely the leg of a bed or desk. I do not see a black backpack in this immediate area.",
      "As the sequence progresses, the camera viewpoint shifts, revealing more of the room. I can see a chair, a desk, a bed, a mini-fridge, and other furniture commonly found in a dorm room. I will broaden my search area to encompass these new parts of the scene.",
      "I systematically scan the visible parts of the room, starting with the area near the desk and bed on the right side of the image sequence. There are items on the desk and the bed, but no black backpack is immediately obvious in these areas. I also check the floor around this desk.",
      "The view continues to expand to the left, showing more of the room, including another chair in the foreground. I shift my focus to the floor area visible in front of this chair and between the furniture items.",
      "Upon examining the floor in front of and slightly to the right of the chair visible in the foreground, particularly in frames starting around the middle of the sequence, I can clearly see a black object. This object has the characteristic shape and appearance of a backpack lying on the floor.",
      "The black backpack is located on the floor in the central part of the room visible, positioned next to the wooden chair that is prominently featured in the middle-to-later frames of the image sequence. It is on the floor near the legs of this chair."
}

\noindent\textbf{Ground Truth Answer:} \textit{On the floor, next to the bed.}
\vspace{1mm} 

\noindent\textbf{Ground Truth Score:} \textit{7}
\vspace{1mm} 

oindent\textbf{Output (\texttt{EQA-RM}):} \textit{As we pan across the room in the images, the initial focus is on the floor near the bed and desk area. The black backpack is not immediately visible in this initial view. However, as the sequence progresses and the camera shifts towards the left, the black object on the floor becomes more prominent. It is situated next to a chair, confirming its location in the room. Therefore, the black backpack is located on the floor between the bed and the chair. The overall structure of the room is consistent with a typical dorm room, with a bed, a desk, and some furniture. The backpack is the only notable object in the floor area visible throughout the sequence, so its presence is easily identifiable. There is no other object on the floor that resembles a backpack, and the backpack itself is distinguishable by its shape and color.<critique> The backpack is clearly visible in the room, located next to a chair on the floor. The presence of the backpack is a clear and distinct object in the scene, easily identifiable through its shape and color. There is no other object in the floor area that matches the description of a backpack, and the backpack is the only significant object present in the scene. The location of the backpack is straightforward, as it is clearly situated between the bed and the chair.</critique><score>7</score> The backpack is clearly visible in the scene, identifiable by its shape and color. The location is straightforward and the object is prominent enough to be easily identified. There are no other objects in the floor area that are similar, and the backpack is the only item of interest. Therefore, the image provides sufficient information to confidently identify the backpack, making the response a correct deduction. The visual grounding is accurate, and the object is clearly located in the scene. The overall quality of the image is adequate for identifying the backpack. Thus, the score is based on the observed visual evidence and the clarity of the object's identification.}
\vspace{1mm}

\noindent\textbf{Score:} \textit{7}

\end{tcolorbox}
\end{table*}

\newpage
\begin{table*}[t!] 
\centering
\caption{\small{An example EQA case study illustrating \texttt{EQA-RM}'s evaluation of an agent's response, reasoning, and grounding across a sequence of 5 observed frames.}}
\label{tab:eqa_case_study_example_2} 

\begin{tcolorbox}[
    colback=gray!5, 
    colframe=black,   
    width=\linewidth, 
    arc=1.5mm, auto outer arc, 
    left=2mm, right=2mm, top=2mm, bottom=2mm, 
    boxrule=0.9pt,    
    colbacktitle=black!75!white, 
    coltitle=white,              
    fonttitle=\bfseries\sffamily\small, 
    title={EQA Case Study Example: Agent Trajectory Evaluation} 
]

\noindent 
\begin{minipage}{\linewidth} 
\centering 
\includegraphics[width=0.18\textwidth, height=2.5cm, keepaspectratio]{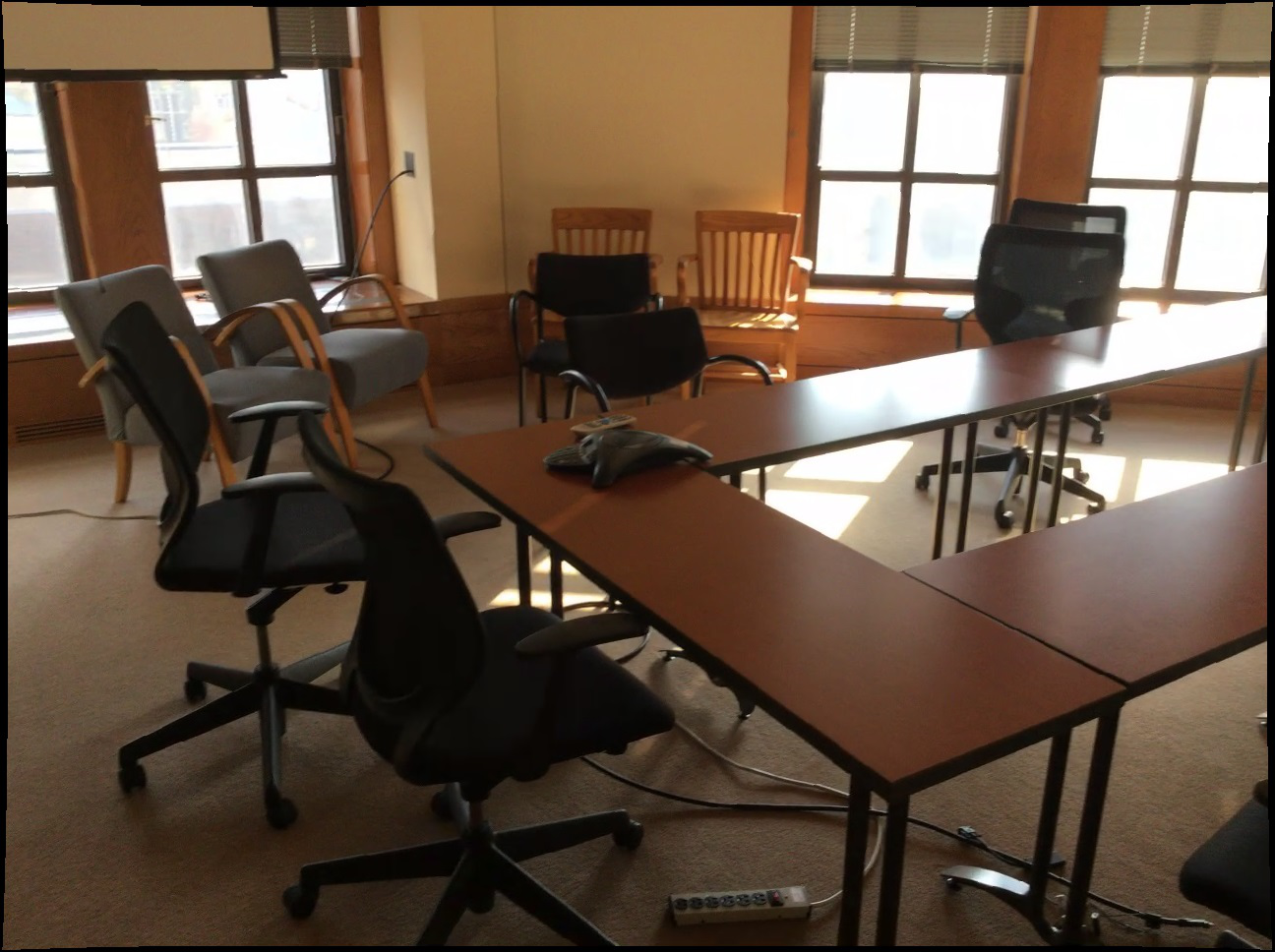}\hfill
\includegraphics[width=0.18\textwidth, height=2.5cm, keepaspectratio]{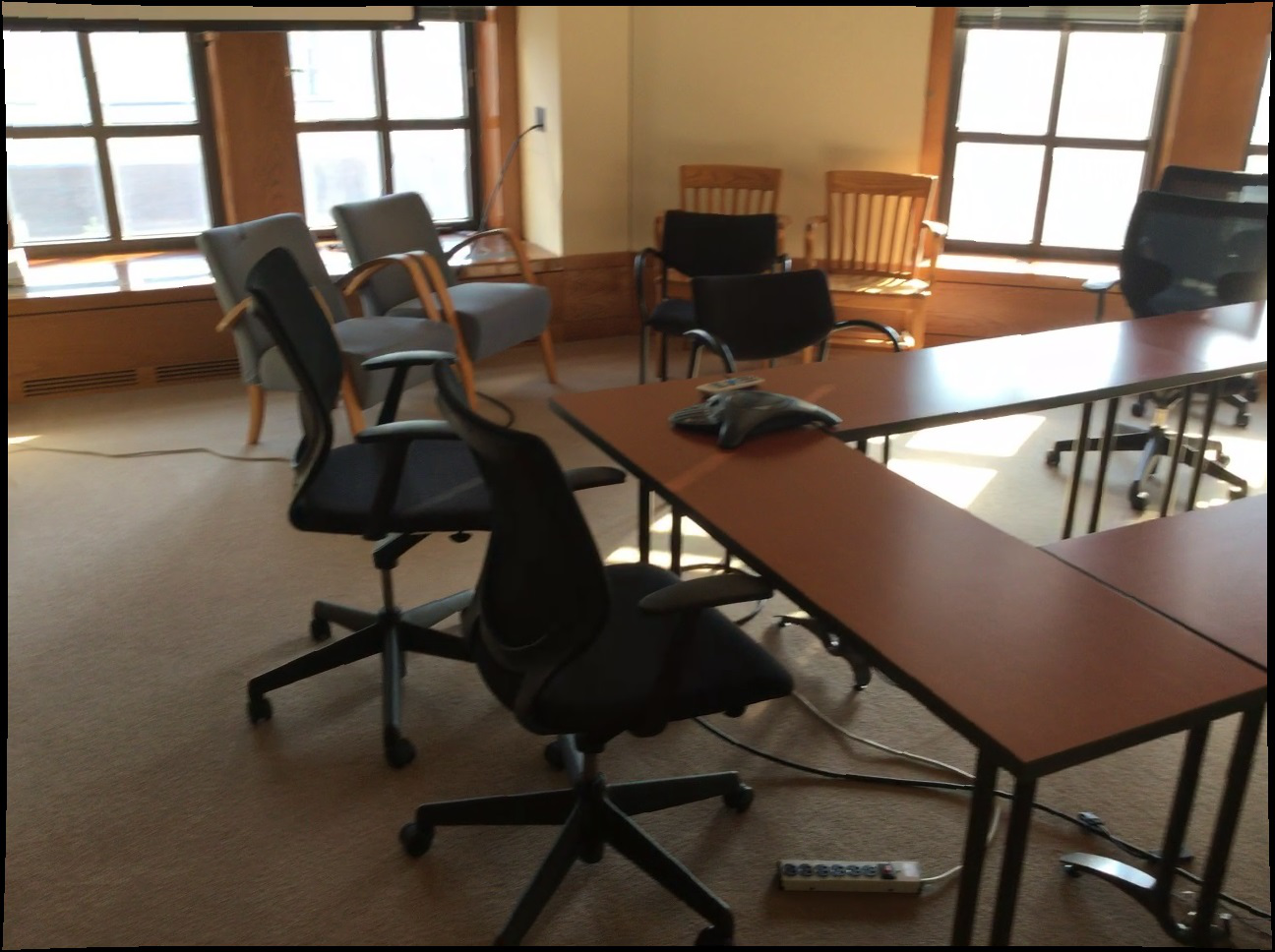}\hfill
\includegraphics[width=0.18\textwidth, height=2.5cm, keepaspectratio]{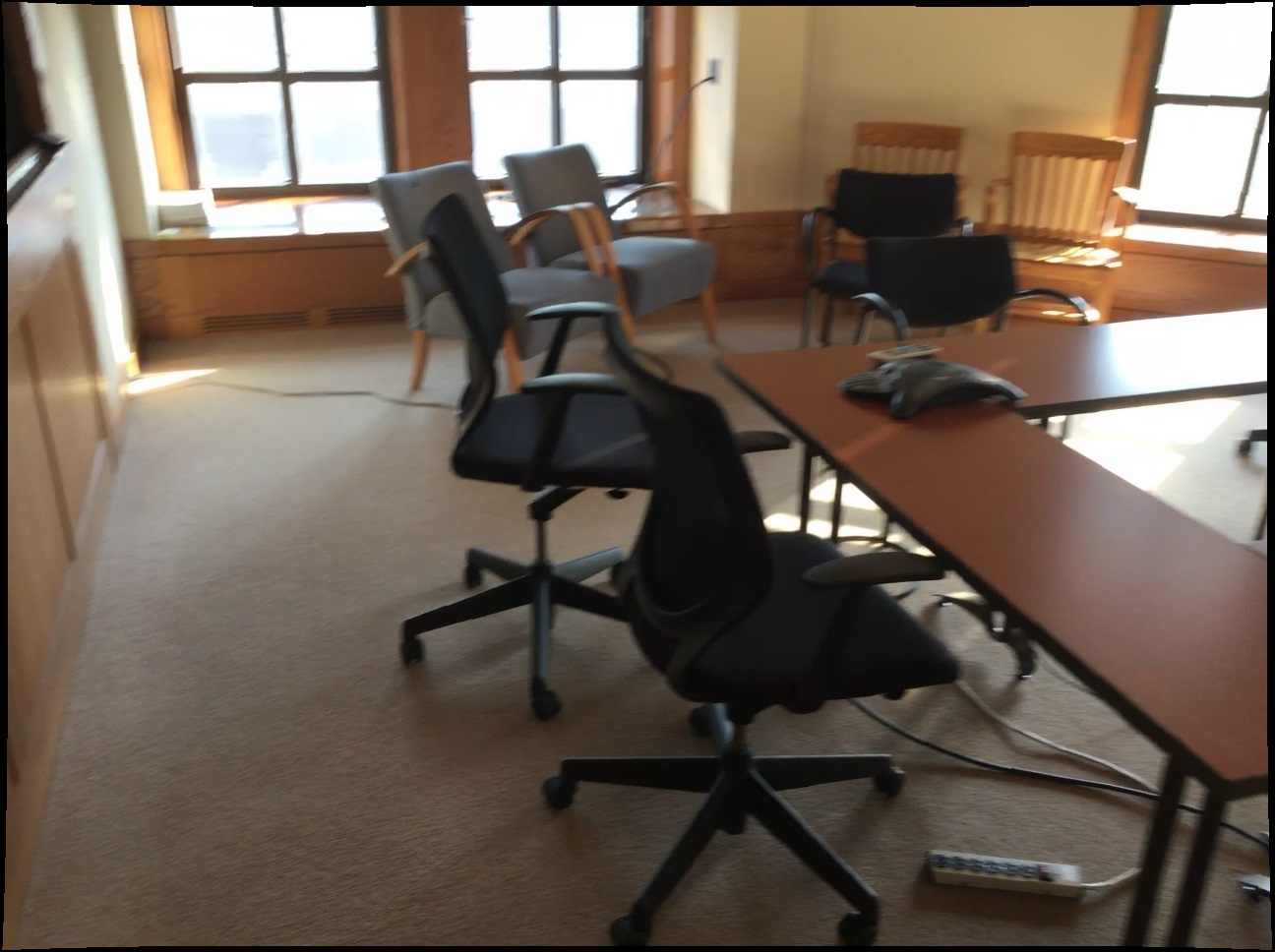}\hfill
\includegraphics[width=0.18\textwidth, height=2.5cm, keepaspectratio]{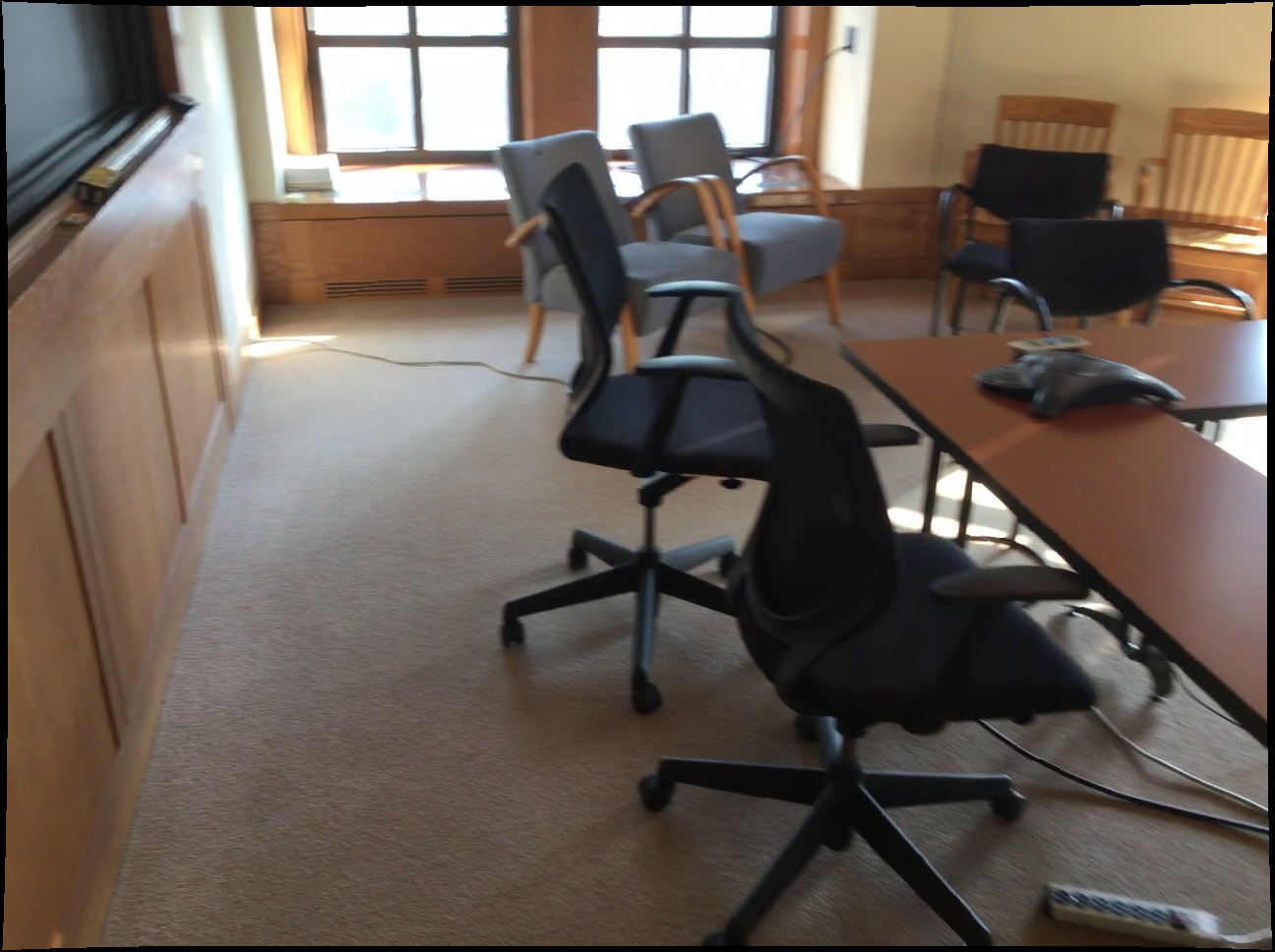}\hfill
\includegraphics[width=0.18\textwidth, height=2.5cm, keepaspectratio]{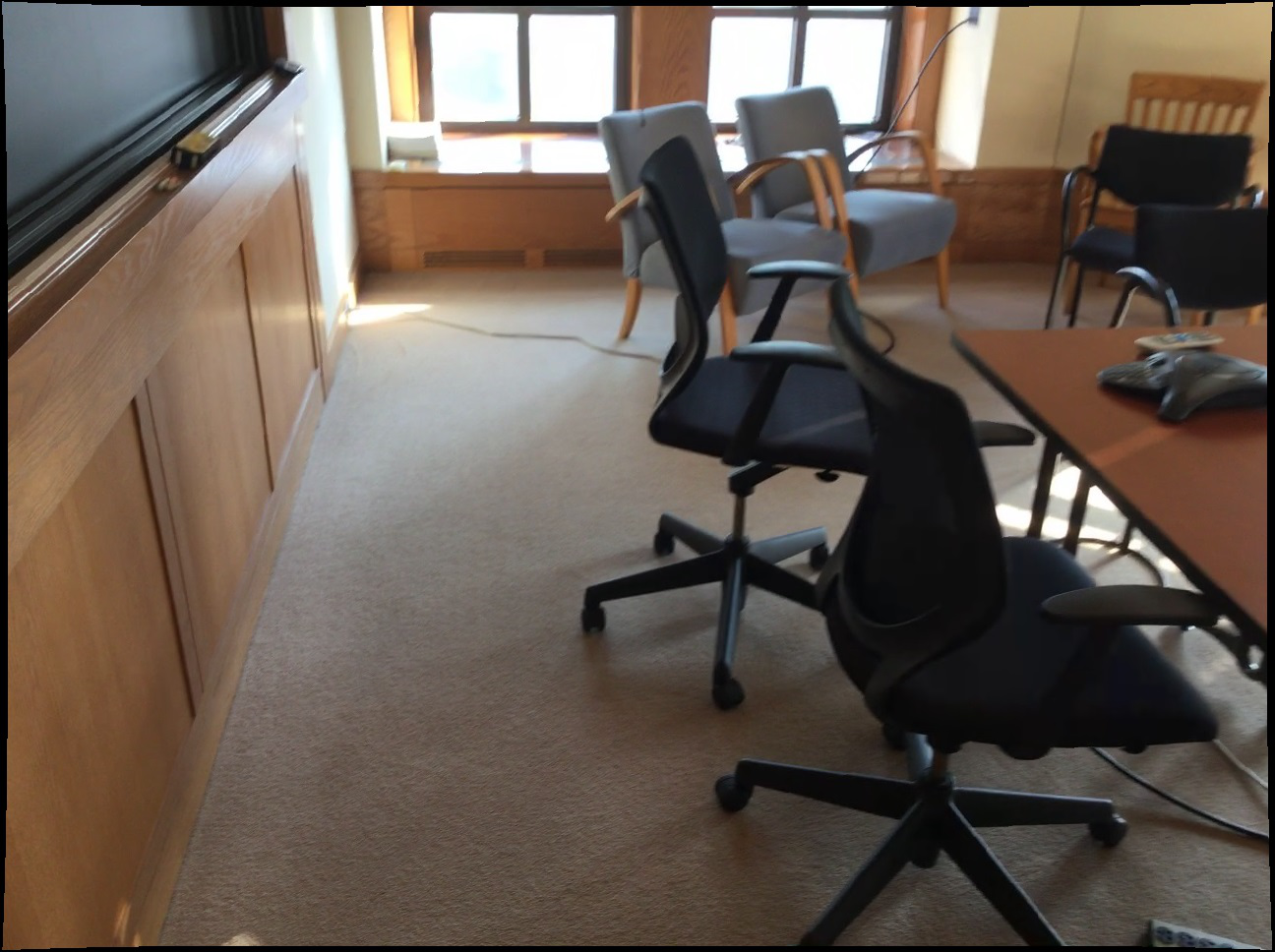}
\end{minipage}

\noindent\textbf{Input:}
\vspace{1mm}

\noindent\textbf{Question:} \textit{Where can I charge my computer?}
\vspace{1mm} 

\noindent\textbf{Predicted Answer:} \textit{You can charge your computer using the power strip located on the floor near the meeting table.}
\vspace{1mm} 

\noindent\textbf{Reasoning Trace:} \textit{"The user is asking about locations to charge a computer within the room shown in the images. I need to examine the provided image sequence to identify electrical outlets or power strips. I will start by looking at the general layout and then scan areas where power sources are commonly found, such as walls, near the floor, and around tables.",
      "In the initial frames, I see a meeting room with a long table, various chairs, windows, a blackboard, and a projector screen. Sunlight is coming through the windows. I'll start by looking near the windows and the floor, as these are common places for outlets or for power strips connecting devices on the table to the wall.",
      "As I move through the frames, the camera pans across the room, showing more of the table, chairs, and windows. I'm specifically looking for wall plates with sockets or any visible power strips or extension cords.",
      "In several frames, particularly those showing the area near the table and chairs, I can see cables running along the floor. Focusing on these cables, in one of the later frames, I clearly see a white power strip with multiple outlets lying on the carpet near the corner of the table and a chair with wheels.",
      "This power strip is on the floor, making it accessible for charging devices used at the table. It appears to have several available outlets. This is a definitive location where a computer could be charged.",
      "Based on the visual evidence from the frames, a power strip was identified on the floor near the table. This provides a charging location."
}

\noindent\textbf{Ground Truth Answer:} \textit{Plug it into the power strip on the floor.}
\vspace{1mm} 

\noindent\textbf{Ground Truth Score:} \textit{10}
\vspace{1mm} 

\noindent\textbf{Output (\texttt{EQA-RM}):} \textit{The user asked where they could charge their computer in the provided room, and the agent identified a power strip on the floor near the table. This is a common location for charging devices. The critique focuses on the logical reasoning and the practical identification of the charging location in the images. The provided answer is correct and directly relevant to the user's query. Therefore, the overall critique is positive, and the score is 10.<critique>While the agent's response is correct and logical, it could be more concise. It's always a good idea to provide a clear and direct answer. The description of the power strip on the floor being near the table and chair wheels is clear, but it could be better to describe it as a power strip or power outlet located on the floor near the table and chair. This makes it more specific and easier to locate. However, the core information is there, and the overall answer is good.</critique><score>10</score>The agent's performance in correctly identifying a charging location within the room is commendable, and the response is concise and clear, fulfilling the user's request. The slight improvement in wording is minor. Overall, a well-crafted response that meets the query's requirements.
}
\vspace{1mm}

\noindent\textbf{Score:} \textit{10}

\end{tcolorbox}
\end{table*}

\end{document}